\documentclass{article}

\usepackage{arxiv}
\usepackage[authoryear, sort]{natbib}
\usepackage[utf8]{inputenc} 
\usepackage[T1]{fontenc}    
\usepackage{hyperref}       
\usepackage{url}            
\usepackage{booktabs}       
\usepackage{amsfonts}       
\usepackage{nicefrac}       
\usepackage{amsmath}
\usepackage{lipsum}
\usepackage{graphicx}
\usepackage{float}
\usepackage{algorithm}  
\usepackage{algpseudocode}
\usepackage{color}
\usepackage{setspace}
\usepackage{arydshln}

\graphicspath{ {./} }

\title{Diffusion models for multivariate subsurface generation and efficient probabilistic inversion}

\author{
 Roberto Miele \\
  Institute of Earth Sciences\\
  University of Lausanne \\
  Lausanne, Switzerland\\
  \texttt{roberto.miele@unil.ch} \\
   \And
 Niklas Linde \\
  Institute of Earth Sciences\\
  University of Lausanne \\
  Lausanne, Switzerland\\
  \texttt{niklas.linde@unil.ch} \\
}

\begin{document}

\maketitle
\begin{abstract}
Diffusion models offer stable training and state-of-the-art performance for deep generative modeling tasks. Here, we consider their use in the context of multivariate subsurface modeling and probabilistic inversion. We first demonstrate that diffusion models enhance multivariate modeling capabilities compared to variational autoencoders and generative adversarial networks. In diffusion modeling, the generative process involves a comparatively large number of generative steps with update rules that can be modified to account for conditioning data. We propose different corrections to the popular Diffusion Posterior Sampling approach by Chung et al. (2023). In particular, we introduce a likelihood approximation accounting for the noise-contamination that is inherent in diffusion modeling. We assess performance in a multivariate geological scenario involving facies and correlated acoustic impedances. Conditional modeling is demonstrated using both local hard data (well logs) and nonlinear geophysics (fullstack seismic data). Our tests show significantly improved statistical robustness, enhanced sampling of the posterior probability density function and reduced computational costs, compared to the original approach. The method can be used with both hard and indirect conditioning data, individually or simultaneously. As the inversion is included within the diffusion process, it is faster than other methods requiring an outer-loop around the generative model, such as Markov chain Monte Carlo.
\end{abstract}


\section{Introduction}\label{Intro}

Predicting the spatial distribution of subsurface properties is a key challenge in the Earth and environmental sciences. Probabilistic inverse modeling \citep{granaSeismicReservoirModeling2021, rubinHydrogeophysics2005} aims to approximate the posterior probability density function (pdf) by integrating prior pdfs with observed data in the form of a likelihood function \citep{tarantolaInverseProblemTheory2005}. Conditioning data may consist of local measurements (e.g., well logs) or indirect geophysical data describing averaged subsurface properties through linear or nonlinear physical forward operators. Prior pdfs expressed as explicit statistical models (e.g., two-point geostatistical models) are computationally convenient, yet are often gross simplifications of geological heterogeneity, leading to biased predictions and incomplete uncertainty quantification \citep[e.g.,][]{lindeUncertaintyQuantificationHydrogeology2017, gomez-hernandezBeNotBe1998, neuweilerEstimationEffectiveParameters2011, journelEntropySpatialDisorder1993, zhouComparisonSequentialIndicator2018}. Alternatively, prior geological knowledge can be represented using 2D or 3D gridded data, defined as training images (TI) that implicitly describe lower- and higher-order spatial relationships. Multiple-Point Statistics (MPS) simulation methods \citep{guardianoMultivariateGeostatisticsBivariate1993,strebelleConditionalSimulationComplex2002,mariethozDirectSamplingMethod2010,mariethozMultiplePointGeostatisticsStochastic2014} rely on TIs to draw geologically-realistic model realizations and have found wide applicability in both hydro-geology \citep[e.g.,][]{hoyerMultiplepointStatisticalSimulation2017, lecozUseMultiplepointStatistics2017, straubhaarMultiplepointStatisticsUsing2020} and reservoir characterization \citep[e.g.,][\& references therein]{melnikovaHistoryMatchingSmooth2015, caersMultiplepointGeostatisticsQuantitative2004}. These realizations can easily be made to honor local observations \citep[][]{straubhaarConditioningMultiplepointStatistics2016, hansenMultiplePointStatistical2018, straubhaarConditioningMultiplePointStatistics2021}, but struggle with high computational costs and poor posterior exploration when used in the context of nonlinear inversion \citep[e.g.,][]{zahnerImageSynthesisGraph2016, levyConditioningMultiplepointStatistics2024, laloyMergingParallelTempering2016, hansenInverseProblemsNontrivial2012, mariethozBayesianInverseProblem2010}. 

The substantial progress of deep generative models over the past decade \citep{goodfellowDeepLearning2016} have created a strong interest in leveraging such techniques in the geosciences. Initial work considered convolution-based deep neural networks, such as variational autoencoders \citep[VAE; ][]{kingmaAutoEncodingVariationalBayes2013} and generative adversarial networks \citep[GAN; ][]{goodfellowGenerativeAdversarialNetworks2014}. These methods can encode complex geological patterns of a TI within a low-dimensional manifold, defining prior pdfs \citep{laloyInversionUsingNew2017, laloyTrainingImageBasedGeostatistical2018}. Inversion can then be performed within the latent space of the pre-trained networks \citep{laloyInversionUsingNew2017, laloyTrainingImageBasedGeostatistical2018, oliviero-durmusGenerativeModellingMeets} using sampling methods, such as Markov chain Monte Carlo \citep[MCMC; ][]{laloyInversionUsingNew2017, laloyTrainingImageBasedGeostatistical2018, levyUsingDeepGenerative2021, mosserStochasticSeismicWaveform2020, liuUncertaintyQuantificationStochastic2022} and Sequential Monte Carlo \citep[SMC; ][]{doucetSequentialMonteCarlo2001, amayaAdaptiveSequentialMonte2021}, ensemble-based approaches \citep{canchumuniRobustParameterizationConditioning2019, liuPetrophysicalCharacterizationDeep2020, moIntegrationAdversarialAutoencoders2020}, deterministic optimization \citep{laloyGradientbasedDeterministicInversion2019, dupontGeneratingRealisticGeology2018}, or variational inference  \citep{chanParametricGenerationConditional2019, levyVariationalBayesianInference2023, mieleDeepGenerativeNetworks2024}. Contrarily to MPS algorithms, conditioning to local data remains cumbersome. One can train the generative models as inference methods on specific conditional data, but this necessitates retraining for each application task \citep[e.g.][]{mosserConditioningGenerativeAdversarial2018, zhangUnetGenerativeAdversarial2021, laloyApproachingGeoscientificInverse2020,fengStochasticFaciesInversion2024, mielePhysicsinformedWNetGAN2024}. Moreover, GANs are prone to unstable training and mode collapse \citep[e.g., ][]{lucicAreGANsCreated2018a}, and the non-linearity of the generative process often adversely affects the inversion results \citep{lopez-alvisDeepGenerativeModels2021,laloyGradientbasedDeterministicInversion2019}. Formulations based on VAEs are more stable, but offer rather lossy representations \citep[e.g.,][]{levyVariationalBayesianInference2023,mieleDeepGenerativeNetworks2024}.

Diffusion models \citep[DM; ][]{hoDenoisingDiffusionProbabilistic2020,sohl-dicksteinDeepUnsupervisedLearning2015} represent the current state-of-the-art in deep generative modeling, outperforming GANs in terms of accuracy without being affected by mode collapse or unstable training \citep[e.g., ][]{dhariwalDiffusionModelsBeat2021}. These methods are based on reversing a \textit{data corruption} process transforming samples of a target distribution (the TI) into those of a known distribution (typically Gaussian). The generative process involves iteratively denoising samples originating from the latter, with the denoising trajectory being learned by a deep neural network \citep[e.g., ][]{songDenoisingDiffusionImplicit2022, hoDenoisingDiffusionProbabilistic2020,songScoreBasedGenerativeModeling2021,karrasElucidatingDesignSpace2022}. 

In geological subsurface modeling, diffusion models (DMs) have been explored in different contexts. \citet{mosserDeepDiffusionModels2023} showed that autoregressive DMs \citep[][]{hoogeboomAutoregressiveDiffusionModels2022} share structural similarities with sequential geostatistical modeling and demonstrated their potential for simulating multi-scale 2D synthetic channel patterns.  \citet{xuDiffSimDenoisingDiffusion2024} found high modeling accuracy of DMs in analogous geological scenarios and proposed a conditional generative approach, where sparse point observations were incorporated as input channels during training.
\citet{aouf3DClayMicrostructure2025} demonstrated that DMs can be used for 2D and 3D modeling of complex porous geological media, further performing conditional modeling on total porosity values, by training the DM to embed such conditioning parameter and associate it to the desired output.
Conditioning on embedded well-log data was further investigated by \citet{leeLatentDiffusionModel2025} for latent diffusion models \citep[LDM; ][]{rombachHighResolutionImageSynthesis2022}, which perform diffusion within the latent space of a pre-trained encoder-decoder network. \citet{ovangerStatisticalStudyLatent2025} extended this work by providing a thorough geostatistical assessment of LDMs against truncated Gaussian random field modeling \citep{matheronConditionalSimulationGeometry1987, mannsethRelationLevelSet2014} for multi-facies scenarios. The authors present overall high modeling performances, but note biases due to the LDM data compression. 
Diffusion models were also recently introduced  to solve inverse modeling problems in the Earth sciences. \citet{wangPriorRegularizedFull2023} used pre-trained DMs as regularizers within a deterministic full waveform inversion (FWI) of seismic data, integrating the FWI’s iterative updates with DM sampling. This framework, designed for subsurface wavespeed predictions, was extended to multiple elastic properties by \citet{taufikLearnedRegularizationsMultiParameter2024}. Both studies demonstrate improved accuracy and more realistic results compared to conventional FWI for synthetic geological scenarios.  \citet{wangControllableSeismicVelocity2024a} further proposed a controllable DM-based FWI, where the model learns the relationship between embedded conditioning data and wavespeed distributions. Finally, \citet{difedericoLatentDiffusionModels2025} show that stochastic inverse modeling can be carried out by combining a pre-trained unconditioned LDM with an ensemble-based inversion method; the authors demonstrated their approach on facies prediction tasks in the context of history matching.

Beyond   Earth sciences, the field of inverse modeling with DMs is of broad interest for computer vision and biomedical imaging, and several methodologies were recently proposed; we refer to \citet{darasSurveyDiffusionModels2024} and \citet{zhaoConditionalSamplingGenerative} for comprehensive overviews. Analogously to methods based on GANs and VAEs, direct training on conditioning data is a popular approach \citep[e.g.,][]{hoClassifierFreeDiffusionGuidance2021, dhariwalDiffusionModelsBeat2021, whangDeblurringStochasticRefinement2022,liuI$^2$SBImagetoImageSchrodinger2023, xuDiffSimDenoisingDiffusion2024, leeLatentDiffusionModel2025, aouf3DClayMicrostructure2025, wangControllableSeismicVelocity2024a}, but implies computationally demanding re-training when considering new data types or designs. Alternatively, one can rely on inference frameworks that iterate \textit{outside} the generative process of pre-trained unconditional DMs. Conventional sampling algorithms (e.g., MCMC) are often inefficient in this setting, due to the DM's large dimensionality and comparatively long generation times \citep{zhaoConditionalSamplingGenerative}. More efficient variational inference algorithms have been proposed for posterior approximation with surrogate modeling, but they often struggle to capture complex or multimodal posterior distributions \citep[e.g.,][]{mardaniVariationalPerspectiveSolving2023, alkanVariationalDiffusionModels2023, fengVariationalBayesianImaging2024a}. Ensemble-based approaches \citep[e.g., ][]{difedericoLatentDiffusionModels2025} can be limited by comparatively high computational costs; LDMs can be used to reduce dimensionality, but the associated encoder-decoder networks (e.g., VAE or GAN) introduce additional complications such as reduced modeling ability \citep[e.g., leading to bias or smoothing; ][]{ovangerStatisticalStudyLatent2025} and non-linearities \citep{darasSurveyDiffusionModels2024}.

Another family of conditioning approaches leverage the sequential nature of the denoising process. For example, SMC can be used to  progressively sample, at different denoising steps, a posterior distribution that evolves towards the posterior pdf \citep[e.g., ][]{trippeDiffusionProbabilisticModeling2022, cardosoMonteCarloGuided2023, wuPracticalAsymptoticallyExact2024, douDIFFUSIONPOSTERIORSAMPLING2024}. These asymptotically exact methods work under the assumption of infinite sampling and linear physics. As an alternative, the denoising process itself can be adapted to achieve approximate Bayesian posterior sampling at low computational costs \citep{songScoreBasedGenerativeModeling2021}, provided that the intractable likelihood is approximated at each denoising step. The approximated likelihood score is then combined with the pre-trained denoising score to steer the denoising generative process towards posterior samples. Among numerous such approaches \citep[see e.g., ][]{darasSurveyDiffusionModels2024}, Diffusion Posterior Sampling (DPS) \citep{chung2023diffusion} has gained wide popularity as it allows inference for both linear and nonlinear inverse problems. Nonetheless, this class of conditional samplers are affected by the likelihood approximation \citep{darasSurveyDiffusionModels2024}. For instance, the accuracy of DPS decreases with the conditioning data noise magnitude and increasing nonlinearity of the problem at hand \citep{chung2023diffusion}. Various adjustments have been proposed in the implementation of DPS, including ad-hoc weighting of the likelihood score. Several studies have highlighted these limitations and proposed partial improvements for linear inverse problems \citep[e.g., ][]{hamidiEnhancingDiffusionModels2025, boysTweedieMomentProjected2024, yismaw2025gaussian, song2023pseudoinverseguided}. 

In this work, we investigate DMs for multivariate subsurface geological modeling and propose improvements to DPS for inversion tasks. Our corrected DPS (CDPS) relies on a likelihood approximation that considers not only the observational error of the conditioning data, but also the impact of the noise contamination at the different stages of the DM. The modified likelihood approximation removes the need to arbitrarily weight the likelihood score and is expected to improve robustness and uncertainty quantification. We further highlight inconsistencies in the original implementation of DPS by \citet{chung2023diffusion} implemented with Denoising Diffusion Probabilistic and Implicit Modeling \citep[DDPM and DDIM; ][]{hoDenoisingDiffusionProbabilistic2020,songDenoisingDiffusionImplicit2022} and suggest corrections.
In our study, we rely on a state-of-the-art DM implementation \citep[EDM; ][]{karrasElucidatingDesignSpace2022} to model a bivariate prior pdf describing sedimentary channel sequences and correlated acoustic impedances. We first demonstrate substantial increases in prior modeling performance against a GAN and a VAE previously introduced for analogous tasks \citep{mieleDeepGenerativeNetworks2024}. Then, we use CDPS to condition realizations on both local observations (well logs) and nonlinear geophysical data (fullstack seismic data), considering Gaussian data noise of different magnitudes. We demonstrate that CDPS yields significantly improved accuracy and enhanced robustness compared to DPS across all considered applications, while allowing for a reduction in the required number of denoising steps. Our results for the linear conditioning case compare well with those sampled with the preconditioned Crank–Nicolson (pCN) MCMC algorithm \citep{cotterMCMCMethodsFunctions2013}. The CDPS can easily be adapted to different DM formulations and data error levels. We further present the algorithm for DDPM and DDIM, with example applications, in the Appendix.

\section{Methodology}\label{Method}
\subsection{Diffusion modeling}\label{General_diff}
Diffusion models enable deep generative modeling by transforming samples from a tractable distribution (often Gaussian) into samples from an unknown target pdf $p(\textbf{x}_0)$ on $\mathbb{R}^d$, from which samples are available (the TI). The diffusion process typically involves corrupting these samples,  $\textbf{x}_0\sim p(\textbf{x}_0)$, through the progressive addition of Gaussian noise at different time-steps $t$, thereby defining the \textit{forward process} $p(\textbf{x}_{t} | \textbf{x}_{t-1})$. After a sufficient time $T$, the corrupted data distribution $p(\textbf{x}_T)$ can be considered Gaussian \citep{hoDenoisingDiffusionProbabilistic2020,sohl-dicksteinDeepUnsupervisedLearning2015}. The \textit{backward process}, $q(\textbf{x}_{0}|\textbf{x}_{T:1})$, represents the sampling or generative process, which is analytically intractable for an arbitrary $p(\textbf{x}_0)$. Numerical methods can be used instead by relying on a trained deep neural network $S_\theta(-)$  \citep{hoDenoisingDiffusionProbabilistic2020,sohl-dicksteinDeepUnsupervisedLearning2015}. 

A generalized continuous-time formulation of diffusion modeling, referred to as \textit{score-based} \citep[e.g.,][]{songScoreBasedGenerativeModeling2021, NEURIPS2019_3001ef25},  represents the forward process through stochastic differential equations (SDE) of the form \citep{songScoreBasedGenerativeModeling2021}
\begin{equation}\label{Eq: SDE_Forward}
    d\textbf{x}_t = f(\textbf{x}_t, t) dt + g(t) d\textbf{W}_t,
\end{equation}

\noindent where $ f(\textbf{x}_t, t)$ is a drift term, $g(t)$ is a diffusion coefficient and $\textbf{W}_t$ is a Wiener process (i.e., Brownian motion). The reverse-time SDE for unconditional sampling is given by \citep{andersonReversetimeDiffusionEquation1982,songScoreBasedGenerativeModeling2021}
\begin{equation}\label{Eq: SDE_Backward}
    dx_t = \left(f(\textbf{x}_t, t) dt + g(t)^2 \nabla_{\textbf{x}_t} \log p_t(\textbf{x}_t)\right)dt + g(t)d\textbf{W}_t,
\end{equation}

\noindent where $\nabla_{\textbf{x}_t} \log p_t(\textbf{x}_t)$ is the score function of the noisy data distribution $\textbf{x}_t$. The generation of samples using Eq. \ref{Eq: SDE_Backward} is possible using Langevin dynamics, implying that a distribution of samples $p(\textbf{x}_0|\textbf{x}_T)$ can be obtained from a single $\textbf{x}_T$. As an alternative, deterministic sampling can be performed using the ordinary differential equation (ODE) \citep{songScoreBasedGenerativeModeling2021}
\begin{equation}\label{Eq: prob_flow_ODE}
    \frac{d\textbf{x}_t}{dt} = \left(f(\textbf{x}_t, t) dt + \frac{g(t)^2}{2} \nabla_{\textbf{x}_t}\log p_t(\textbf{x}_t)\right),
\end{equation}

\noindent which enables a bivariate mapping between $\textbf{x}_T$ and $\textbf{x}_0$, and is referred to as \textit{probability flow ODE}. The diffusion can be simulated within \textit{Variance preserving} (VP) (i.e., the added noise variance is bounded) or \textit{Variance exploding} (VE) frameworks. For the latter, the forward process (Eq. \ref{Eq: SDE_Forward}) at a given time-step $t$ can be defined by 
\begin{equation}\label{Eq: VESDE_forward}
\textbf{x}_t = \textbf{x}_0 + \sigma_t\ z, \qquad z \sim \mathcal{N}(0,I),
\end{equation}

\noindent where $\sigma_t$ represents the time-dependent noise magnitude. Using Tweedie's formula \citep{robbinsEMPIRICALBAYESAPPROACH, efronTweediesFormulaSelection2011}, the score function is
\begin{equation}\label{Eq: Tweedie}
    \nabla_{\textbf{x}_t} \log p_t(\textbf{x}_t) = \frac{\mathbb{E}[\textbf{x}_0|\textbf{x}_t] - \textbf{x}_t}{\sigma_t^2},
\end{equation}

\noindent where $\mathbb{E}[\textbf{x}_0|\textbf{x}_t]$ is a time-dependent conditional expectation of $\textbf{x}_0$ given $\textbf{x}_\textit{t}$.  Neural networks can be trained to parameterize the score function from noisy data and a given time \textit{t}, $S_\theta(\textbf{x}_t,\ t)$, by minimizing the mean square error \citep{songScoreBasedGenerativeModeling2021}
\begin{equation}\label{Eq: loss1}
    \mathcal{L}(S_\theta) = \mathbb{E}_{\textbf{x}_{0}}\mathbb{E}_{\textbf{x}_{t}|\textbf{x}_{0}}\left[||S_\theta(\textbf{x}_t,\ t) - \nabla_{\textbf{x}_t} \log  p_t(\textbf{x}_t|\textbf{x}_0)||_2^2\right]
\end{equation}

\noindent or directly as a denoiser $S_\theta(\textbf{x}_0+\textbf{z},\ t)$, using \citep[e.g.,][]{karrasElucidatingDesignSpace2022}
\begin{equation}\label{Eq: loss2}
    \mathcal{L}(S_\theta) = \mathbb{E}_{\textbf{x}_{0}}\mathbb{E}_{\textbf{z}\sim\mathcal{N}(0,\sigma^2I)}||S_\theta(\textbf{x}_0+\textbf{z},\ t) - \textbf{x}_0)||_2^2.
\end{equation}

\noindent The trained network is finally used in the backward process (Eq. \ref{Eq: SDE_Backward}) for generative (prior) modeling.

The original formulations for DMs are the discrete-time DDPM \citep{hoDenoisingDiffusionProbabilistic2020,sohl-dicksteinDeepUnsupervisedLearning2015} and its generalization DDIM \citep{songDenoisingDiffusionImplicit2022}.  Here, $S_\theta(\textbf{x}_t, t)$ is trained to predict the noise in $\textbf{x}_t$, $\epsilon_{\theta}^{(t)}$, and the generative process for both DDPM and DDIM is defined by
\begin{equation}\label{Eq: DDIM}
    \textbf{x}_{t-1} = \sqrt{\bar{\alpha}_{t-1}}
    \left(\frac{\textbf{x}_t-\sqrt{1-\bar{\alpha}_t}\ \epsilon_{\theta}^{(t)} }{\sqrt{\bar{\alpha}_t}}\right) 
    +\sqrt{1-\bar{\alpha}_{t-1}-\tilde{\sigma}_t^2} \epsilon_{\theta}^{(t)}
    + \tilde{\sigma}_t\epsilon_t,
\end{equation}

\noindent where $\beta_t$ defines the noise schedule of the diffusion process, $\alpha_t = 1-\beta_t$, $\hat{\alpha}_t = \prod_{s=1}^t \alpha_s$ and $\tilde{\sigma}_t^2$ is the conditional variance sequence (injected noise during sampling), and $\epsilon_t\sim \mathcal{N}(0,I)$. DDIM and DDPM are discretizations of score-based diffusion, with $\epsilon^{(t)} = -\nabla_{\textbf{x}_t}\log p_t(\textbf{x}_t) \sqrt{1-\bar{\alpha}_t}$ \citep{songScoreBasedGenerativeModeling2021, dhariwalDiffusionModelsBeat2021}. In DDPM, $\tilde{\sigma}_t = \sqrt{\frac{1-\bar{\alpha}_{t-1}}{1-\bar{\alpha}_t}}\sqrt{\frac{1-\bar{\alpha}_t}{\bar{\alpha}_{t-1}}}$, while in the DDIM framework $\tilde{\sigma}_t$ is set to $0$ for deterministic sampling \citep{hoDenoisingDiffusionProbabilistic2020, songScoreBasedGenerativeModeling2021}, and can hence be associated to a VP SDE and VP ODE, respectively. 

\subsection{Diffusion Posterior Sampling (DPS)}\label{DPS}
In Bayesian inversion, one is interested in characterizing or sampling from the posterior pdf, $p(\textbf{x}_0|\textbf{d})$, with the data vector $\textbf{d}$ given by
\begin{equation}\label{Eq: inversion}
    \textbf{d} = \mathcal{F}(\textbf{x}_0) + \epsilon_\textbf{d},
\end{equation}

\noindent where $\mathcal{F(-)}$ corresponds to a forward operator (either linear or nonlinear) and $\epsilon_\textbf{d}$ is the noise in the observed data. The inversion problem can be formalized by Bayes' theorem, for a constant model dimension, as $p(\textbf{x}_0|\textbf{d}) \propto p(\textbf{d}|\textbf{x}_0)p(\textbf{x}_0)$, where $p(\textbf{d}|\textbf{x}_0)$ is the likelihood function.

Unconditional modeling from $p(\textbf{x}_0)$ can be achieved with a trained DM, using Eqs. \ref{Eq: SDE_Backward} or \ref{Eq: prob_flow_ODE}. Following Bayes' rule, sampling from the posterior pdf can in theory be achieved using the score function \citep{songScoreBasedGenerativeModeling2021}
\begin{equation}\label{Eq: posterior_grad}
    \nabla_{\textbf{x}_t}\log p_t(\textbf{x}_t|\textbf{d}) = \nabla_{\textbf{x}_t}\log p_t(\textbf{x}_t) + \nabla_{\textbf{x}_t}\log p_t(\textbf{d}|\textbf{x}_t), 
\end{equation}

\noindent where the first term can be approximated by the trained neural network $S_\theta(\textbf{x}_t, t)$ and $\nabla_{\textbf{x}_t}\log p_t(\textbf{d}|\textbf{x}_t)$ is a noise-dependent likelihood score. In what follows, we will refer to $\nabla_{\textbf{x}_t}\log p_t(\textbf{x}_t)$ as the \textit{prior score}. If $S_\theta (\textbf{x}_t, t)\simeq \nabla_{\textbf{x}_t} \log p_t (\textbf{x}_t)$ (Eq. \ref{Eq: loss1}), then $\hat{\textbf{x}}_0^{(t)}$ is derived using Tweedie's formula (Eq. \ref{Eq: Tweedie}). Otherwise, it is directly obtained as $\hat{\textbf{x}}_0^{(t)} = S_\theta (\textbf{x}_t, t)$ (Eq. \ref{Eq: loss2}). The likelihood score is given by an intractable integral 
\begin{equation}\label{Eq: intract_likelihood}
    p_t(\textbf{d}|\textbf{x}_t) = \int{p(\textbf{d}|\textbf{x}_0)p(\textbf{x}_0|\textbf{x}_t)d\textbf{x}_0}.
\end{equation}

\citet{chung2023diffusion} note that for a given $\textbf{x}_t$, an estimate of the expected mean for $p(\textbf{x}_0|\textbf{x}_t)$ is given by the denoiser $\hat{\textbf{x}}_0^{(t)} = \mathbb{E}[\textbf{x}_0|\textbf{x}_t]$ (Eq. \ref{Eq: Tweedie}); therefore, the likelihood can be approximated as $p(\textbf{d}|\textbf{x}_t) \simeq p(\textbf{d}|\hat{\textbf{x}}_0^{(t)})$, and
\begin{equation}\label{Eq: xhat_likelihood}
    \nabla_{\textbf{x}_t} \log  p(\textbf{d}|\textbf{x}_t) \simeq \nabla_{\textbf{x}_t} \log p(\textbf{d}|\hat{\textbf{x}}_0^{(t)}).
\end{equation}

\noindent In practice, the score in Eq. \ref{Eq: xhat_likelihood} is computed by backpropagating the log-likelihood approximations at each time step $t$. This approach is referred to as Diffusion Posterior Sampling (DPS), and was proposed by \citet{chung2023diffusion} as a general inference framework for both linear and nonlinear inverse problems. Assuming uncorrelated and homoscedastic Gaussian data errors, \citet{chung2023diffusion} use the likelihood 
\begin{equation}\label{Eq: like_gen_dps}
    p(\textbf{d}|\hat{\textbf{x}}_0^{(t)}) = \frac{1}{\sqrt{(2\pi)^n\sigma_\textbf{d}^{2n}}}\exp{\left(-\frac{||\mathcal{F}(\hat{\textbf{x}}_0^{(t)})-\textbf{d}||_2^2}{2\sigma_\textbf{d}^2}\right)}, 
\end{equation}

\noindent where $\sigma_\textbf{d}$ is the standard deviation of the data noise $\epsilon_\textbf{d}$, leading to 
\begin{equation}\label{Eq: grad_like_gen_dps}
    \nabla_{\textbf{x}_t}\log p(\textbf{d}|\textbf{x}_t) \simeq \nabla_{\textbf{x}_t}\log p(\textbf{d}|\hat{\textbf{x}}_0^{(t)}) = -\frac{1}{\sigma_\textbf{d}^2} \nabla_{\textbf{x}_t} ||\mathcal{F}(\hat{\textbf{x}}_0^{(t)})-\textbf{d}||_2^2.
\end{equation}

This approximation does not account for errors in the $\hat{\textbf{x}}_0^{(t)} estimate$. In their implementation, \citet{chung2023diffusion} propose a weight to the likelihood score, replacing $\frac{1}{\sigma_\textbf{d}^2}$ with a hyperparameter $\rho$, to account for the impact of errors in the likelihood approximation. In their accompanying code, the authors further replace the sum of square error in Eq. \ref{Eq: grad_like_gen_dps} with the root-mean-square error (RMSE), justifying this choice as a solution for increased stability \citep[see open review in][]{chung2023diffusion}. This arbitrary choice will drastically decrease the magnitude of the likelihood score and, hence, decrease the weight given to the data in the sampling process.  

The conditional score approximation used within DPS results in a too low variance of the generated samples; it has been argued that the sampling process is rather associated to the maximization of a posterior than to posterior sampling \citep{xuRethinkingDiffusionPosterior2025}. Improvements to this approximation should include the impact of errors in the denoiser ($\hat{x}_0^{(t)}$) \citep{song2023pseudoinverseguided, hamidiEnhancingDiffusionModels2025, yismaw2025gaussian, boysTweedieMomentProjected2024, pengImprovingDiffusionModels2024}. Following \citet{efronTweediesFormulaSelection2011}, the denoiser's variance can be computed using Tweedie's formula for the second moment
\begin{equation}\label{Eq: TweediesVariance}
    \operatorname{Var}\{\hat{\textbf{x}}_0^{(t)}|\textbf{x}_t\} = \sigma_\textbf{t} (1+\sigma_\textbf{t} \nabla_{\textbf{x}_t}^2 \log p_t (\textbf{x}_t)),
\end{equation}

\noindent where $\nabla_{\textbf{x}_t}^2 \log p_t (\textbf{x}_t)$ is the Hessian matrix of the score of the data distribution $\textbf{x}_t$. Using the Hessian directly is computationally demanding, as its calculation scales exponentially with the model's dimensionality \citep{boysTweedieMomentProjected2024}. The Covariance-Aware Diffusion Posterior Sampling \citep{hamidiEnhancingDiffusionModels2025} for linear inversion uses an approximation of the Hessian, computed as the difference between the scores at consecutive and sufficiently small time-steps (i.e., using the finite difference method). The $\Pi$GDM \citep{song2023pseudoinverseguided} and CoDPS \citep{yismaw2025gaussian} methods use time-step dependent-heuristic approximations of the denoiser's standard deviation for linear inverse problems, representing significantly faster solutions.  \citet{pengImprovingDiffusionModels2024} derived a formulation to derive the denoiser variance ($\tilde{\sigma_t}$ in Eq. \ref{Eq: DDIM}) for specific types of DM networks \citep{dhariwalDiffusionModelsBeat2021}. For more general applications, they suggest to estimate the denoiser variance directly from the denoising errors of the trained model, using known test data. In the following, we adopt the latter approach, and introduce a general and consistent formulation for approximate posterior sampling for both linear and nonlinear conditioning data.

\subsubsection{Corrected Diffusion Posterior Sampling (CDPS)}\label{corrected_DPS}
The proposed corrected DPS (CDPS), in which the likelihood formulation accounts for the denoising-step-dependent error $\epsilon_{\hat{\textbf{x}}_0}^{(t)}$ of the trained model. The $\epsilon_{\hat{\textbf{x}}_0}^{(t)}$ is the largest at early steps of the backward process and decreases with decreasing $\sigma_\textbf{t}$. The forward problem at time \textit{t }can be written as (c.f., Eq. \ref{Eq: inversion}). 
\begin{equation}\label{Eq: newinverse}
    \textbf{d} = \mathcal{F}(\hat{\textbf{x}}_0^{(t)}+\epsilon_{\hat{\textbf{x}}_0}^{(t)}) + \epsilon_\textbf{d}
\end{equation}

Assuming an ideal denoiser, we assume $\epsilon_{\hat{\textbf{x}}_0}^{(t)}$ to be spatially uncorrelated and Gaussian-distributed \citep{song2023pseudoinverseguided, yismaw2025gaussian, pengImprovingDiffusionModels2024}. After training, we evaluate the standard deviation of the error for each data dimension (i.e., cells or pixels), using different test data samples $i$ corrupted with known noise, as $\sigma_{{\hat{x}_0},i}^{(t)}\simeq \sqrt{\sum_{i=1}^N(\hat{x}_{0,i}^{(t)}-x_{0,i})^2/N}$. The error distribution depends on the spatial features of the input data, which is computable only through Hessian estimation \citep{efronTweediesFormulaSelection2011,boysTweedieMomentProjected2024,hamidiEnhancingDiffusionModels2025} or using a network predicting conditional covariance \citep[][see Section \ref{DPS}]{pengImprovingDiffusionModels2024}. We assume error homoscedasticity, using the mean value of $\sigma_{{\hat{x}_0}}^{(t)}$ from all the pixels.

In agreement with \citet{lindeUncertaintyQuantificationHydrogeology2017} and \citet{friedliSolvingGeophysicalInversion2024} who studied approximations to intractable likelihoods arising in geophysics, we propagate the errors, $\sigma_{{\hat{x}_0},i}^{(t)}$, through the forward model using a first-order Taylor expansion $\textbf{x} \mapsto \mathcal{F(\textbf{x})}$ around $\hat{\textbf{x}}_0^{(t)}$, 
\begin{equation}\label{Eq: Linearinverse}
    \textbf{d} \approx \mathcal{F}(\hat{\textbf{x}}_0^{(t)}) + \mathcal{J}_{\hat{\textbf{x}}_0^{(t)}} \epsilon_{\hat{\textbf{x}}_0}^{(t)} + \epsilon_\textbf{d},
\end{equation}

\noindent where $\mathcal{J}_{\hat{\textbf{x}}_0^{(t)}}$ is the corresponding Jacobian matrix of the forward operator. Assuming Gaussian distributions $p(\epsilon_{\hat{\textbf{x}}_0}^{(t)}) = \mathcal{N}(0,\sigma_{\hat{\textbf{x}}_0}^{2^{(t)}}\textbf{I}) = \mathcal{N}(0,\Sigma_{\hat{\textbf{x}}_0}^{(t)}) $ and $p(\epsilon_\textbf{d}) = \mathcal{N}(0,\sigma_\textbf{d}^2\textbf{I}) = \mathcal{N}(0,\Sigma_\textbf{d})$, the likelihood $p(\textbf{d}|\hat{\textbf{x}}_0^{(t)})$ can be approximated as
\begin{equation}\label{Eq: nonlin_like}
    p(\textbf{d}|\textbf{x}_t) \simeq p(\textbf{d}|\hat{\textbf{x}}_0^{(t)}) = \mathcal{N}(0,\tilde{\Sigma}_\textbf{d}^{(t)}),
\end{equation}

\noindent  where $\tilde{\Sigma}_\textbf{d}^{(t)} = \mathcal{J}_{\hat{\textbf{x}}_0^{(t)}}^T \Sigma_{\hat{\textbf{x}}_0}^{(t)} \mathcal{J}_{\hat{\textbf{x}}_0^{(t)}} + \Sigma_\textbf{d}$ (see \citet{friedliSolvingGeophysicalInversion2024} for further details). For a linear operator (e.g., local conditioning), the full covariance matrix in Eq. \ref{Eq: nonlin_like} is obtained without any approximations, using matrix multiplication with the linear operator. Finally, we re-write the likelihood score (Eq. \ref{Eq: grad_like_gen_dps}) as 
\begin{equation}\label{Eq: fixed_gradient_likelihood}
    \nabla_{\textbf{x}_t} \log  p(\textbf{d}|\textbf{x}_t) \simeq \nabla_{\textbf{x}_t} \log  p(\textbf{d}|\hat{\textbf{x}}_0^{(t)}) = -\nabla_{\textbf{x}_t} \left[(\mathcal{F}(\hat{\textbf{x}}_0^{(t)})-\textbf{d})^T \tilde{\Sigma}_\textbf{d}^{-1^{(t)}} (\mathcal{F}(\hat{\textbf{x}}_0^{(t)})-\textbf{d})\right].
\end{equation}

\noindent This is the main adaptation introduced by CDPS. 

We report on another error in the original DPS implementation on DDPM \citep{chung2023diffusion}. At each step, the authors use the likelihood score to obtain the conditional denoised image as $\textbf{x}_{t-1} = \textbf{x}_{t-1}' - \rho \nabla_{\textbf{x}_t} \log p(\textbf{d}|\hat{\textbf{x}}_0^{(t)})$, where $\textbf{x}_{t-1}'$ is the denoised image considering the prior score, obtained with Eq. \ref{Eq: DDIM}  as is. However, this is inconsistent with Bayes' theorem \citep[e.g., Eq. \ref{Eq: posterior_grad}; ][]{songScoreBasedGenerativeModeling2021}, according to which Eq. \ref{Eq: DDIM} in the conditioning case should be
\begin{equation}\label{Eq: DDIM_DPS}
    \textbf{x}_{t-1} = \sqrt{\bar{\alpha}_{t-1}}
    \left(\frac{\textbf{x}_t-\sqrt{1-\bar{\alpha}_t}\ \hat{  \epsilon}_{\theta}^{(t)} }{\sqrt{\bar{\alpha}_t}}\right) 
    +\sqrt{1-\bar{\alpha}_{t-1}-\sigma_t^2}\hat{  \epsilon}_{\theta}^{(t)}
    + \sigma_t\epsilon_t.
\end{equation}

\noindent where $\hat{  \epsilon}_{\theta}^{(t)}=\epsilon_{\theta}^{(t)}-\sqrt{1-\bar{\alpha}_t}\nabla_{\textbf{x}_t} \log p(\textbf{d}|\hat{\textbf{x}}_0^{(t)})$, following the guided diffusion modeling approach proposed by \citet{dhariwalDiffusionModelsBeat2021}. In practice, even if one does not consider the error in $\hat{\textbf{x}}_0^{(t)}$, the likelihood score $\nabla_{\textbf{x}_t} \log p(\textbf{d}|\hat{\textbf{x}}_0^{(t)})$ should be rescaled to the diffusion noise magnitudes rather than to a fixed arbitrary $\rho$. We suggest that this error may be an additional cause of instability of the DPS method; we provide an implementation of the CDPS for DDPM and DDIM in Appendix \ref{AppC}, implementing this correction. The CDPS algorithm for score-based diffusion is given in Algorithm \ref{alg:Alg1}, specifically within the EDM diffusion framework \citep{karrasElucidatingDesignSpace2022}. Further details on EDM are provided in Section \ref{modeling}.

\begin{algorithm}
    \caption{Corrected DPS (CDPS) algorithm within the EDM diffusion modeling framework by \citet{karrasElucidatingDesignSpace2022}}\label{alg:Alg1}
    \begin{algorithmic}[1] 
    
    \State \textbf{Require:} trained network $S_\theta(-)$, conditioning data $\textbf{d}$, forward operator $\mathcal{F}$, maximum number of time steps $N$, noise schedule $\sigma_i$, limits $\sigma_{max}$ and $\sigma_{min}$
    \Statex
    
    \For{ $\sigma_t = 0,..., \sigma_{max}$}
        \State \textbf{sample} $\textbf{x}_0 \sim q_{data}$ \Comment{Evaluate denoiser uncertainty for the sampling noise levels} 
        \State $\textbf{x}_t \gets \textbf{x}_0 + \sigma_\textbf{t}\textbf{z}, \textbf{z} \sim \mathcal{N}(0,I)$
        \State $\hat{\textbf{x}}_0^{(t)} \gets S_\theta(\textbf{x}_t, t)$ 
        \State $\sigma_{{\hat{x}_0}}^{(t)}\gets \mathbb{E}\left[\sqrt{\sum_{i=1}^N(\hat{x}_{0,i}^{(t)}-x_{0,i})^2/N}\right]$
    \EndFor
    \Statex
    \State \textbf{sample} $\textbf{x}_T \sim \mathcal{N}(0,I)$
    \For{\textit{i = 0,..., N}}
        \State $\sigma_t \gets \sigma_i$ \Comment{Get $\sigma_t$ according to the sampling schedule (Eq. \ref{Eq: karras_sampling_schedule}) } 
        \State $\hat{\textbf{x}}_0^{(t)} \gets S_\theta(\textbf{x}_t, \sigma_t)$ \Comment{Evaluate the denoised image at noise level $\sigma_t$}
        \State $\nabla_{x_t} \log p_t (\textbf{x}_t) \gets (S_\theta(\textbf{x}_t, \sigma_t) - \textbf{x}_t) / \sigma_t^2$ \Comment{Evaluate denoising score at $\sigma_t$}
        \Statex
        \State $\Sigma_{\hat{\textbf{x}}_0}^{(t)} \gets \sigma_{\hat{\textbf{x}}_0}^{2^{(t)}}$  \Comment{Get denoiser uncertainty at $t$}
    \If{$\mathcal{F}$ is nonlinear} \Comment{Propagate denoiser uncertainty in the \textbf{d} domain}
        \State $\mathcal{J}_{\hat{\textbf{x}}_0} \gets \mathcal{J}_\mathcal{F} (\hat{\textbf{x}}_0^{(t)})$
        \State $\tilde{\Sigma}_\textbf{d} \gets \mathcal{J}_{\hat{\textbf{x}}_0^{(t)}}^T \Sigma_{\hat{\textbf{x}}_0}^{(t)} \mathcal{J}_{\hat{\textbf{x}}_0^{(t)}} + \Sigma_\textbf{d}$
    \Else
        \State $\tilde{\Sigma}_\textbf{d} \gets \mathcal{F}^T \Sigma_{\hat{\textbf{x}}_0}^{(t)} \mathcal{F} + \Sigma_\textbf{d}$
    \EndIf   
    \Statex
    \State $\nabla_{x_t} \log p_t (\textbf{d}|\hat{\textbf{x}}_0^{(t)}) \gets -\nabla_{\textbf{x}_t} \left[(\mathcal{F}(\hat{\textbf{x}}_0^{(t)})-\textbf{d})^T \tilde{\Sigma}_\textbf{d}^{-1^{(t)}} (\mathcal{F}(\hat{\textbf{x}}_0^{(t)})-\textbf{d})\right]$ \Comment{Evaluate likelihood score given $\textbf{x}_0^{(t)}$}
    \State $\nabla_{x_t} \log p_t (\hat{\textbf{x}}_0^{(t)}|\textbf{d}) \gets \nabla_{x_t} \log p_t (\textbf{d}|\hat{\textbf{x}}_0^{(t)}) + \nabla_{x_t} \log p_t (\textbf{x}_t)$ \Comment{Evaluate posterior score}
    \State $d\textbf{x}_t\gets  -\sigma_t \nabla_{x_t} \log p_t (\hat{\textbf{x}}_0^{(t)}|\textbf{d})$ \Comment{Evaluate $d\textbf{x}/dt$  at $\sigma_t$}
    \State $\textbf{x}_{t-1} \gets x_t + (\sigma_{t-1} - \sigma_t) d\textbf{x}_t$  \Comment{Take Euler step from $\sigma_t$ to $\sigma_{t-1}$}
    \Statex
    \If{$\sigma_{t-1} \not= 0$} \Comment{Apply $2^{\text{nd}}$ order correction}
        \State $\hat{\textbf{x}}_0^{(t-1)} \gets S_\theta(\textbf{x}_{t-1}, \sigma_{t-1})$
        \State $\nabla_{x_{t-1}} \log p_{t-1} (\textbf{x}_{t-1})' \gets (S_\theta(\textbf{x}_{t-1}, \sigma_{t-1}) - \textbf{x}_{t-1}) / \sigma_{{t-1}}^2$
        \State $\tilde{\Sigma}_\textbf{d}^{-1^{(t-1)}} \gets \text{Repeat lines 13--19 for } {(t-1)}$
        \State $\nabla_{x_{t-1}} \log p_{t-1} (\textbf{d}|\hat{\textbf{x}}_0^{(t-1)})'  \gets  -\nabla_{\textbf{x}_{t-1}} \left[(\mathcal{F}(\hat{\textbf{x}}_0^{(t-1)})-\textbf{d})^T \tilde{\Sigma}_\textbf{d}^{-1^{(t-1)}} (\mathcal{F}(\hat{\textbf{x}}_0^{(t-1)})-\textbf{d})\right]$
        \State $\nabla_{x_{t-1}} \log p_{t-1} (\hat{\textbf{x}}_0^{({t-1})}|\textbf{d})' = \nabla_{x_{t-1}} \log p_{t-1} (\textbf{d}|\hat{\textbf{x}}_0^{({t-1})})' + \nabla_{x_{t-1}} \log p_{t-1} (\textbf{x}_{t-1})'$
        \State $d\textbf{x}_t' \gets  -\sigma_{t-1} \nabla_{x_{t-1}} \log p_{t-1} (\hat{\textbf{x}}_0^{({t-1})}|\textbf{d})'$
        \State $x_{t-1} \gets  x_t + (\sigma_{t} - \sigma_{t-1})(\frac{1}{2}d\textbf{x}_t+\frac{1}{2}d\textbf{x}_t')$

    \EndIf
\EndFor
\State \textbf{Output:} $\tilde{\textbf{x}}_0 \sim p(\textbf{x}_0|\textbf{d}) $
\end{algorithmic} 
\end{algorithm} 

\subsection{Multivariate diffusion modeling}\label{modeling}

The simultaneous sampling of multivariate subsurface properties is carried out within a single generative process. We do not consider formulations involving dimensionality reduction with LDMs \citep{rombachHighResolutionImageSynthesis2022}. We adopt a VE score-based DM framework, often referred to as EDM \citep{karrasElucidatingDesignSpace2022}, and use its probability flow ODE for unconditional modeling \citep{karrasElucidatingDesignSpace2022}:
\begin{equation}\label{Eq: karras_ode}
    d\textbf{x} = -\frac{\sigma_t}{dt} \sigma_t \nabla_{\textbf{x}_t} \log p_t (\textbf{x}_t).
\end{equation}

For posterior sampling (Eq. \ref{Eq: posterior_grad}), we substitute the score with $\nabla_{\textbf{x}_t}\log p_t(\textbf{x}_t|\textbf{d})$ using our proposed likelihood approximation (Eq. \ref{Eq: fixed_gradient_likelihood}). Sampling is performed by evaluating the ODE at noise levels given by the schedule \citep{karrasElucidatingDesignSpace2022}
\begin{equation}\label{Eq: karras_sampling_schedule}
    \sigma_{i>0} = \left(\sigma_{max}^{1/\rho} + \frac{i}{N-1}(\sigma_{min}^{1/\rho} - \sigma_{max}^{1/\rho})\right)^{\rho} \quad and \quad \sigma_N = 0,
\end{equation}

\noindent where $\sigma_{max}$ and $\sigma_{min}$ are the maximum and minimum noise levels in the corrupted data, $i$ is the step number, $i \in \{1,...,N\}$,  and $\rho$ is an exponential factor. In agreement with the authors' original choices, we found that $\sigma_{max} = 80$, $\sigma_{min} = 0.002$ and $\rho = 7$ provided the best performance. The motivation behind these parameter choices is explained by \citet{karrasElucidatingDesignSpace2022}. The ODE is evaluated using the Heun’s $2^{\mathrm{nd}}$ order method \citep{ascherComputerMethodsOrdinary1998} (lines 23-36 in Algorithm \ref{alg:Alg1}).

The network $S_\theta(\textbf{x}_t, t)$ is trained as a denoiser, using an improved training algorithm with noise-specific weights integrated in Eq. \ref{Eq: loss2} as proposed by \citet{karrasElucidatingDesignSpace2022}. The network architecture relies on the encoder-decoder structure of a U-Net \citep{ronnebergerUNetConvolutionalNetworks2015}, as implemented by \citet{dhariwalDiffusionModelsBeat2021} for general image generation. We use a multi-head self-attention mechanism \citep{vaswaniAttentionAllYou} at each of its three layers to capture global dependencies. In our implementation, each subsurface property type is represented by a separate channel. Categorical variables such as geological facies assume a single numerical variable within a single dimension. Similarly to \citet{mieleDeepGenerativeNetworks2024} and \citet{mielePhysicsinformedWNetGAN2024}, the multivariate prior distribution $p(\textbf{x}_0)$ is represented by a training dataset where each sample is a joint realization  of two or more co-located subsurface properties. The data can be represented as a single large image (randomly sampled during training) \citep[see, e.g.,][]{laloyTrainingImageBasedGeostatistical2018} or as separate samples. 

%
\subsection{Performance assessment}\label{metrics}
Prior modeling performance is assessed by comparing the unconditional subsurface realizations against the TI and realizations generated with two benchmark models, a GAN and a VAE, proposed by \citet{mieleDeepGenerativeNetworks2024} for an analogous case study. The statistical distance between the target and modeled distributions (either prior or posterior) were quantified using the root-mean-square-error (RMSE) and the Kullback-Leibler divergence ($KL$) \citep{kullbackInformationSufficiency1951}.  The similarity of spatial features in the realizations was assessed using the structural similarity index (SSIM) \citep{wangImageQualityAssessment2004} defined as 
\begin{equation}\label{Eq: SSIM}
    \mathrm{SSIM}(\textbf{u},\textbf{v}) = \frac{(2\mu_\textbf{u} \mu_\textbf{v}+C_1)(2\sigma_{\textbf{uv}}+C_2)}{(\mu_\textbf{u}^2+ \mu_\textbf{v}^2+C_1)(\sigma_\textbf{u}^2+ \sigma_\textbf{v}^2+C_2)},
\end{equation}

\noindent where $\textbf{u}$ and $\textbf{v}$ define sliding windows of size $M \times M$, on two different images, $\mu$ and $\sigma$ are their corresponding mean and standard deviations, and $C$ is a constant. Following the original implementation \citep{wangImageQualityAssessment2004} and previous related works \citep[e.g.,][]{levyVariationalBayesianInference2023, mieleDeepGenerativeNetworks2024}, we set $M = 7$, $C_1 = 0.01$ and $C_2 = 0.03$. The SSIM ranges between -1 and 1, with 1 indicating a perfect match between images. Further metrics concerning the reproduction of facies volume fractions, two-points statistics (i.e., variograms), and lateral continuity (shape) of geological features, are presented in Appendix \ref{AppA}.

Convergence of the inversions results was assessed using the weighted RMSE (WRMSE), a data misfit weighted by the data standard deviation $\sigma_{\textbf{d}}$, defined as
\begin{equation}\label{Eq: WRSME}
    \text{WRMSE}=\sqrt{\frac{1}{N_\textbf{d}}\sum{\left[\frac{\textbf{d}_i-\mathcal{F}(\textbf{x}_{0})_i}{\sigma_{\textbf{d},i}}\right]}^2}.
\end{equation}
A \text{WRMSE} $\leq 1.1$ is chosen to indicate convergence in terms of data misfit as the residuals have a similar statistic as the observational noise and a WRMSE $< 1$ may indicate overfitting of the noisy data. Moreover, the predictive power of the inferred pdf is assessed using the logarithmic scoring rule (logS) \citep{goodRationalDecisions1952}, computed as logS$(\hat{p},\textbf{d}) = -\log \hat{p}(\textbf{d})$, where $\hat{p}$ is the inferred pdf and $\textbf{d}$ is the real data. A lower value of logS corresponds to a better approximation.
The proposed CDPS is compared to the original DPS \citep{chung2023diffusion} and to a sampling method based on Markov chain Monte Carlo (MCMC). For MCMC, we use the preconditioned Crank–Nicolson algorithm (pCN) \citep{cotterMCMCMethodsFunctions2013}, which is well-suited for high-dimensional problems. In this method, each new proposal $\textbf{x}_{new}$ is defined by $
\textbf{x}_{new} = \sqrt{1-\beta^2} \textbf{x}_{old} + \beta \textbf{z}$, where $\textbf{x}_{old}$ represents the previous state, $\beta$ is a constant determining the step size, $\textbf{z}\sim \mathcal{N}(0,\textbf{C})$, and $\textbf{C}$ is the prior covariance, determining the jump rate. The acceptance probability $\alpha$ takes the form $\alpha (\textbf{x}_{new},\textbf{x}_{old})=\min(1,(p(\textbf{d}|\textbf{x}_{new})/p(\textbf{d}|\textbf{x}_{old})))$. Convergence of the pCN samples is determined with the Gelman-Rubin diagnostic \citep{gelmanInferenceIterativeSimulation1992} using $\hat{R} \leq 1.2$ for all model parameters as a criterion. We further assess the exploration of the parameter space by computing within-chain autocorrelation of samples, that is measuring the covariance between two samples divided by the parameter's variance.

\subsection{Code implementation}\label{codeimpl}
The methods were implemented in Python (Python 3.12) using PyTorch (PyTorch 2.5) libraries with CUDA implementation. The training of all the generative models and the corresponding unconditional and conditional modeling were performed using a single GPU NVIDIA\textsuperscript{\texttrademark} GeForce\textsuperscript{\textregistered} GTX TITAN X (CUDA 12.2). The pCN sampling was performed on the same machine, generating DM samples using this GPU. No parallelization on different GPUs was considered. Training, modeling and sampling times mentioned in Section \ref{Res_3} are specific to this machine.

\section{Results}\label{Res}
\begin{figure}
    \centering
    \includegraphics[width=1\linewidth]{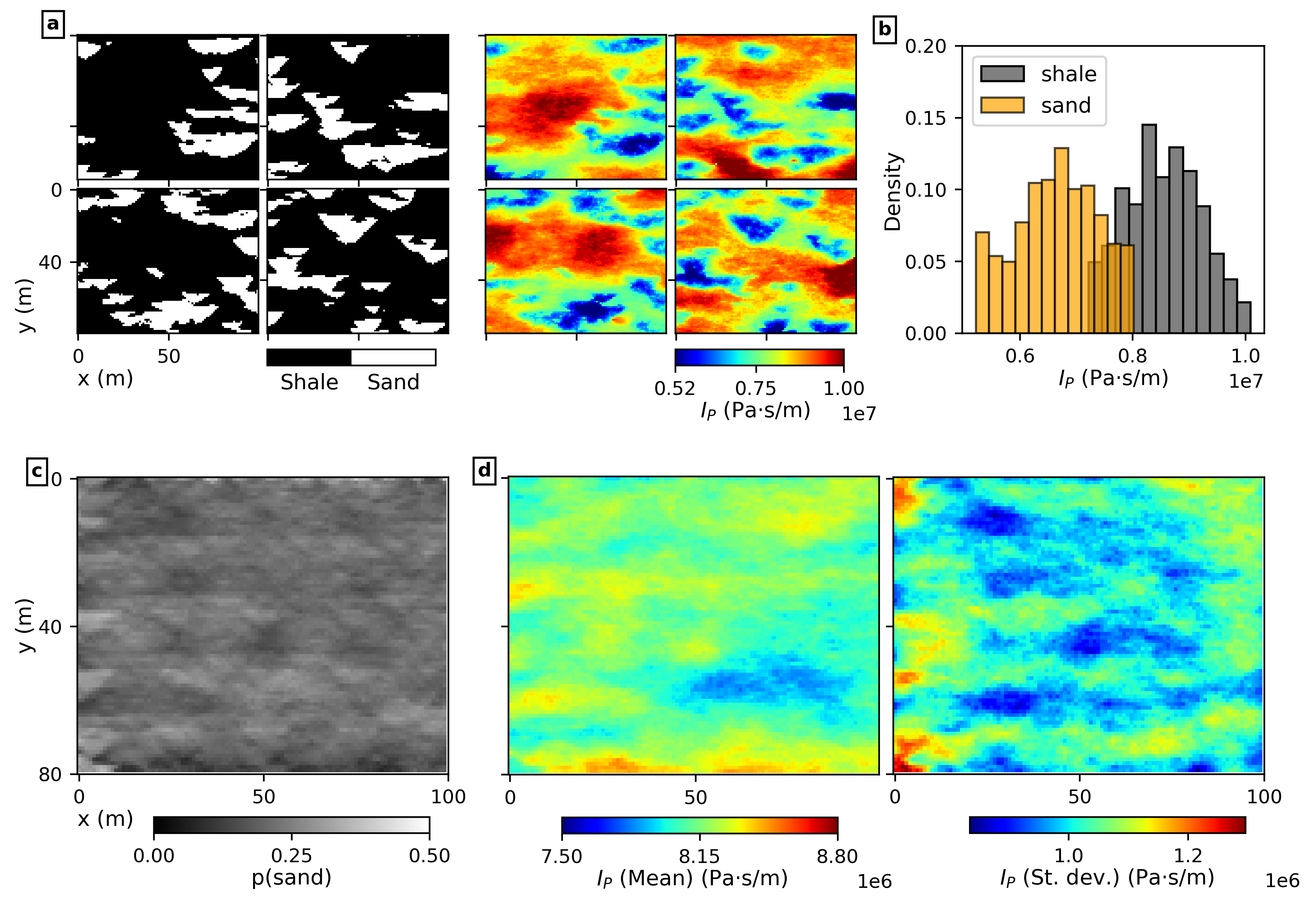}
    \caption{Prior distribution as represented in the training dataset: (a) Facies and $I_P$ realizations; (b) $I_P$ distribution per facies; (c) probability of sand distribution; (d) point-wise average and standard deviation of $I_P$.}
    \label{fig:prior_fig}
\end{figure}

We evaluate the generative performance of the trained DM by parameterizing a bivariate prior pdf, describing a sequence of lenticular sandy channels in a shale background (facies), and the associated acoustic impedances ($I_P$). We consider an area of 100 $\times$ 80 cells, assumed to have sides of 1 m (Fig. \ref{fig:prior_fig}). The TI consists of 3000 samples (Fig. \ref{fig:prior_fig}a) of MPS facies realizations, generated using Petrel software (SLB) based on a 3D conceptual geological model. Shales and sands are categorized as 0 and 1, respectively. The facies distribution ensemble is spatially uniform, with each cell having average \textit{p}(sand) = 0.3  (Fig. \ref{fig:prior_fig}c). The facies-dependent $I_P$ is represented using direct sequential co-simulations with multi-local distribution functions \citep{nunesGeostatisticalSeismicInversion2017, soaresDirectSequentialSimulation2001}. We account for facies-dependent $I_P$ marginal distributions (Fig. \ref{fig:prior_fig}b) and spatial models (2D variograms; detailed in Appendix \ref{AppA}). The average spatial distribution of $I_P$ (Fig. \ref{fig:prior_fig}d) is primarily dependent on facies with an area of relatively lower values in the lower-right section of the considered area. The DM was trained for a total of 600 epochs (225000 steps). The GAN and VAE used for benchmarking were trained on the same TI, for a total of 2000 epochs, using the hyperparameters proposed by \citet{mieleDeepGenerativeNetworks2024}.

\subsection{Unconditonal modeling}\label{Res_1}

\begin{figure}
    \centering
    \includegraphics[width=0.9\linewidth]{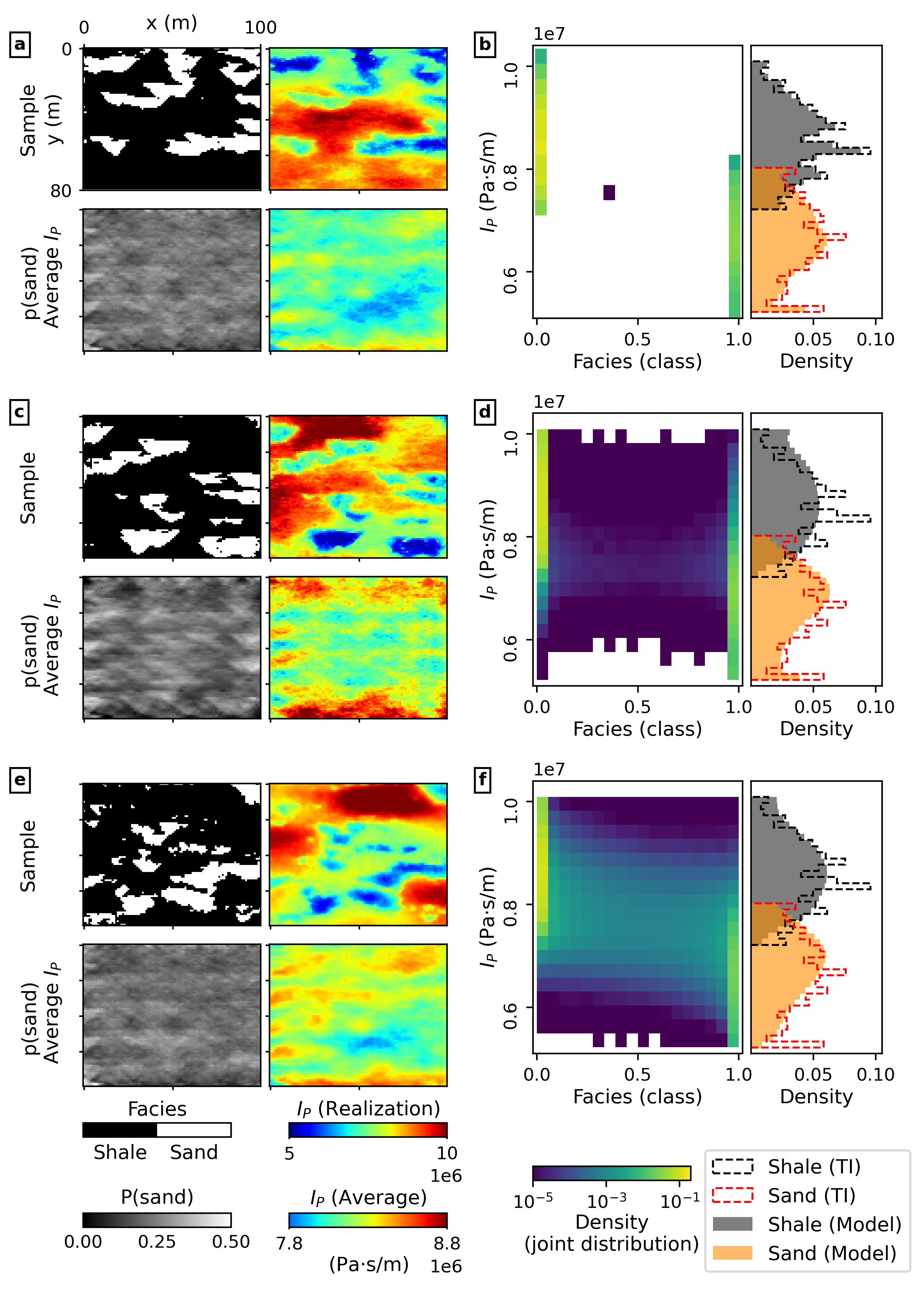}
    \caption{Summary of unconditional modeling performances of (a, b) DM, (c, d) GAN, and  (e, f) VAE. Subplots (a), (c), (e): realizations and average distribution of facies and $I_P$ for (a) DM, (c) GAN and (e) VAE; subplots (b), (d), (f): joint and $I_P$ distributions for (b) DM, (d) GAN and (f) VAE.}
    \label{fig:mod_res_1}
\end{figure}

Prior modeling performances are assessed on 1000 unconditional samples generated from the trained DM, GAN and VAE each. The channel morphologies and $I_P$ spatial patterns reproduced by DM and GAN (\textit{Samples} in Fig. \ref{fig:mod_res_1}a and c) are visually indistinguishable from those of the TI (Fig. \ref{fig:prior_fig}a). For DM, the \textit{p}(sand) and average $I_P$ match the prior accurately; Table \ref{tab:Table1} shows the average relative error for the two property types. 
Of the two benchmarking models, the GAN's \textit{p}(sand) and average $I_P$ distributions (Fig. \ref{fig:mod_res_1}c) show significant deviations from the prior. In fact, the overestimation of \textit{p}(sand) in the central area may indicate mode collapse. Despite the smooth spatial features in the generated samples (Fig. \ref{fig:mod_res_1}e), the VAE performs better than GAN at capturing both properties average distributions, and displays even a larger accuracy than DM for facies (Table \ref{tab:Table1}).

The joint distributions of facies and co-located $I_P$ obtained from the sampled realizations (Fig. \ref{fig:mod_res_1}b, d, and f) show significantly superior modeling ability of the DM (Fig. \ref{fig:mod_res_1}b). The facies classes are modeled with negligible errors (i.e., facies values different from 0 and 1); the corresponding $I_P$ marginal distributions (plotted along the y-axes) match accurately the prior means and standard deviations, fitting the data with a $KL$ divergence approximately five times lower than for GAN and VAE (Table \ref{tab:Table1}). Further analysis of the networks unconditional modeling performances focusing on geostatistical metrics, are provided in Appendix \ref{AppA}.

\begin{table}
\centering
\caption{Metrics for unconditional prior modeling using the trained DM and benchmarking GAN and VAE. The \textit{p}(sand) and Avg. $I_P$ refer to the average models obtained from the realizations ensembles, shown in Figure \ref{fig:mod_res_1}a, c, e. $KL$: Kullback–Leibler divergence; \textit{Avg.}: average; \textit{St. dev.}: standard deviation.}
\label{tab:Table1}
\begin{tabular}{ c | c || c || c c }
\hline
     &  & \textit{p}(sand) \textemdash Avg. $I_P$ & \multicolumn{2}{c}{$I_P$ marginal distributions}\\ 
\hline
     & Lithology & Relative error & \begin{tabular}{@{}c@{}} {Avg.\textpm St. dev.} \\ {($10^6 \, \mathrm{Pa \cdot s / m}$)} \end{tabular}  &  \begin{tabular}{@{}c@{}} {$KL_{I_P}$}\\{($10^{-3}$)} \end{tabular}\\ 
\midrule
Prior & \begin{tabular}{@{}c@{}} \textit{Sand} \\ \textit{Shale} \end{tabular} & - & \begin{tabular}{@{}c@{}} $6.66 \pm 0.73$ \\ $8.54 \pm  0.66$ \end{tabular} & - \\
\hline
DM    & \begin{tabular}{@{}c@{}} \textit{Sand} \\ \textit{Shale} \end{tabular} & 
        \begin{tabular}{@{}c@{}} $8.3\%$ \\ $0.8\%$ \end{tabular} & 
        \begin{tabular}{@{}c@{}} $6.65 \pm 0.73$ \\ $8.53 \pm  0.68$ \end{tabular} & 
        \begin{tabular}{@{}c@{}} $0.25$ \\ $0.42$ \end{tabular} \\
\hline
GAN   & \begin{tabular}{@{}c@{}} \textit{Sand} \\ \textit{Shale} \end{tabular} & 
        \begin{tabular}{@{}c@{}} $21.0\%$ \\ $2.1\%$ \end{tabular} &
        \begin{tabular}{@{}c@{}} $6.75 \pm  0.77$ \\ $8.70 \pm 0.72$\end{tabular} & \begin{tabular}{@{}c@{}} $2.1$ \\ $1.1$ \end{tabular} \\
\hline
VAE  & \begin{tabular}{@{}c@{}} \textit{Sand} \\ \textit{Shale} \end{tabular} & 
        \begin{tabular}{@{}c@{}} $7.0\%$ \\ $1.2\%$ \end{tabular} &
        \begin{tabular}{@{}c@{}} $6.87 \pm  0.76$\\ $8.62 \pm 0.69$\end{tabular} & 
        \begin{tabular}{@{}c@{}} $1.3$ \\ $1.5$ \end{tabular} \\
\hline
\end{tabular} 
\end{table}

\subsection{Inverse modeling using CDPS}\label{Res_2}
\begin{figure}
    \centering
    \includegraphics[width=.9\linewidth]{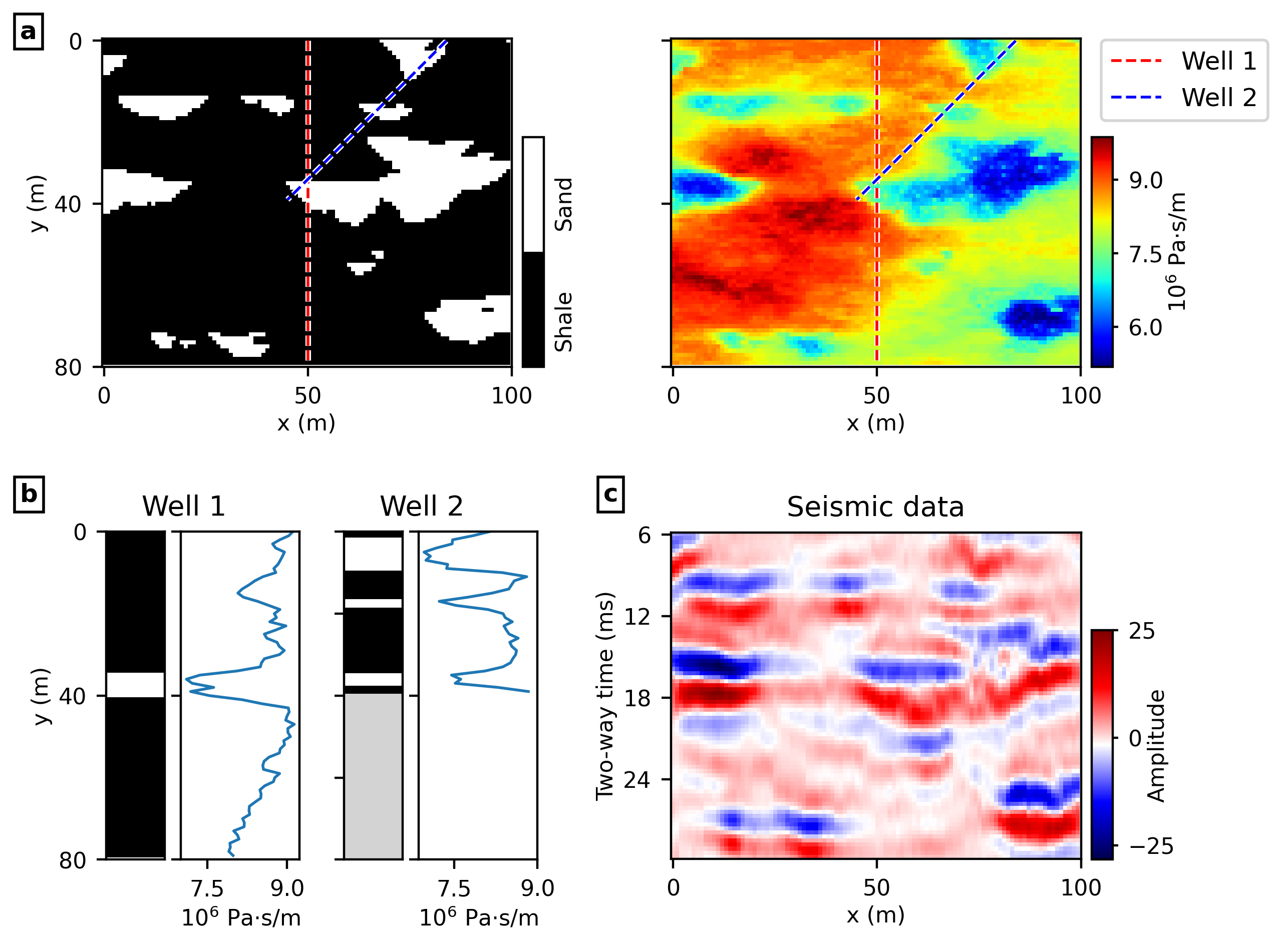}
    \caption{Test model used in inverse modeling: (a) real subsurface distribution ("True") and location of the two wells used for the linear inversion; (b) conditioning well log data for the linear inversion; (c) conditioning seismic data for the nonlinear inversion.}
    \label{fig:inv_test_data}
\end{figure}

We evaluated the proposed CDPS by considering both linear and nonlinear inverse problems using a test realization from the prior distribution as our "True" facies and $I_P$ (Fig. \ref{fig:inv_test_data}a). The linear test case consists in generating samples that are locally conditioned on direct observations. The data, $\bf{d_{obs}}$, are obtained from two wells ("Well 1" and "Well 2" in Figure \ref{fig:inv_test_data}a). This problem is analogous to an \textit{image inpainting} task in computer vision, where missing parts of images are generated conditioned on the known regions.
The nonlinear example is a geophysical inversion problem with fullstack seismic data (Fig. \ref{fig:inv_test_data}c), assumed to be derived from the true $I_P$, and modeled as follows. First, vertical sequences of impedance contrasts ($R$) are computed following $R_i = (I_{P_{i+1}} -I_{P_{i}}) / (I_{P_{i+1}} +I_{P_{i}})$, where the subscript $i$ indicates an $I_P$ sample. Then, each trace is convolved with a known seismic source wavelet; in our case, a Ricker wavelet with central frequency of 25Hz and resolution of 3 ms. For both case studies, we evaluated the CDPS under different Gaussian data noise conditions (Table \ref{tab:Table2}). 

We compare the results to those obtained with DPS \citep{chung2023diffusion}. For fair comparison of the two theoretical approaches, we implemented the DPS within the EDM framework used in this work: following DPS theory, we compute the actual Gaussian log-likelihood considering $\bf{d_{obs}}$ noise only (e.g., Eq. \ref{Eq: grad_like_gen_dps} for absolute $\sigma_\textbf{d}$) and sum it to the prior score for denoising (i.e., Algorithm \ref{alg:Alg1} without lines 13-19).

\begin{figure}
    \centering
    \includegraphics[width=.9\linewidth]{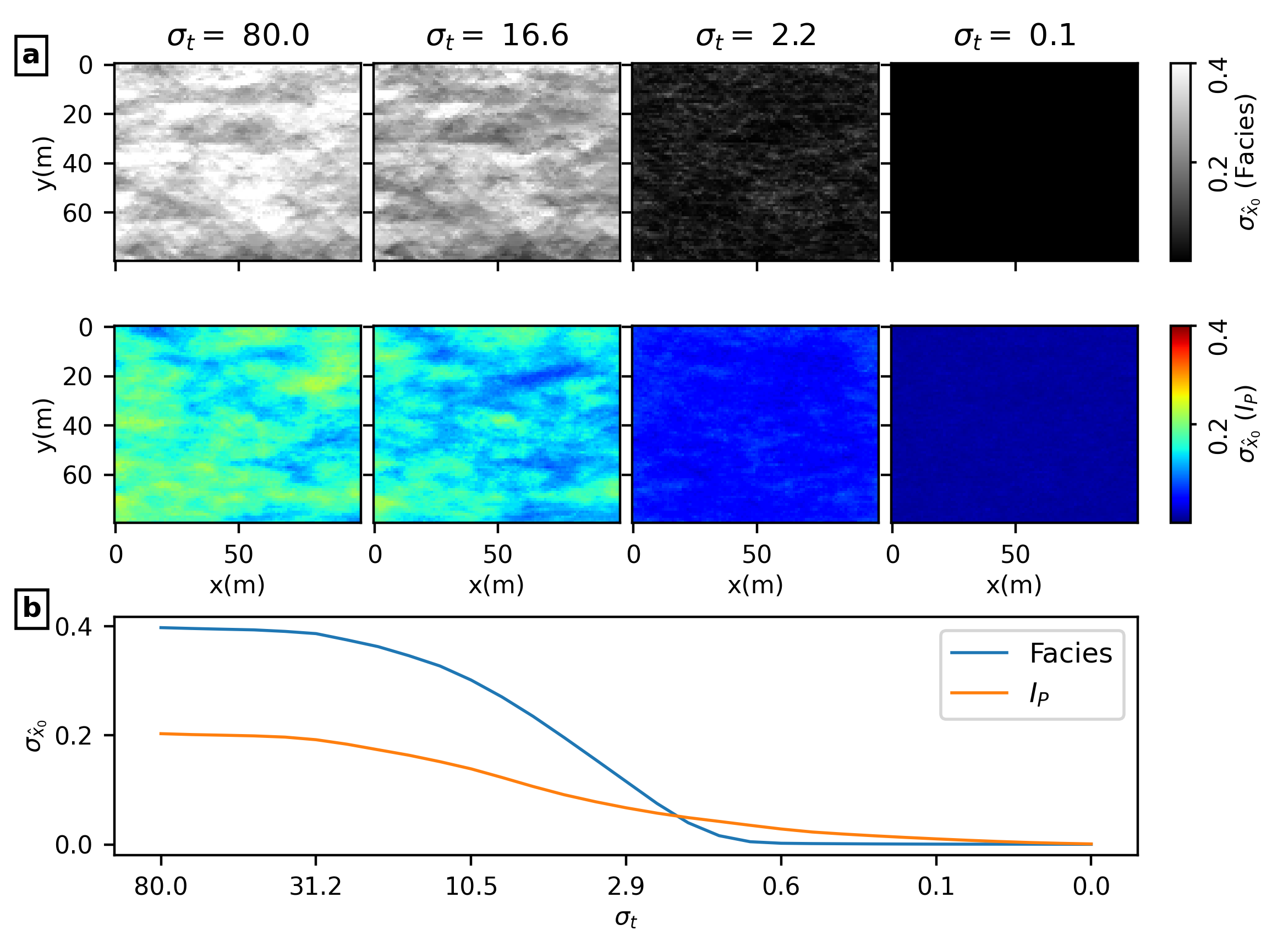}
    \caption{Modeling error: (a) examples of errors at specific noise levels;  (b) standard deviation of the homoscedastic error considered. The values refer to the raw network output, of range $[0,1]$.}
    \label{fig:err_sigmax0}
\end{figure}

Using 300 validation examples from the TI we evaluated the modeling error $\sigma_{\hat{\textbf{x}}_0}^{(t)}$ for 1000 denoising steps. Examples of pixel-dependent uncertainties are shown in Fig. \ref{fig:err_sigmax0}a, while the full homoscedastic noise sequence used is shown in Fig. \ref{fig:err_sigmax0}b. Here, even if the facies classes are categorical, they are modeled as a continuous variable by DM and treated as such in the likelihood score computation. 
In Appendix \ref{AppB}, we present applications of CDPS using well and seismic data simultaneously.

\begin{table}
\centering
\caption{Noise contamination levels of the conditioning data ($\bf{d_{obs}}$) for the linear and nonlinear inversion examples.}
\label{tab:Table2}
\begin{tabular}{ c | c | c }
\hline
& \begin{tabular}{@{}c@{}} Linear \\ $\sigma_{Fac.}$, $\sigma_{I_{P}}$ ($10^6 \, \mathrm{Pa \cdot s / m}$) \end{tabular} & \begin{tabular}{@{}c@{}} Nonlinear \\ $r$, $\sigma_c$ (Amplitude) \end{tabular}\\ 
\hline
Data noise 1 & $0.05, 0.12$ & $0.025, 0.5$ \\
\hline
Data noise 2 & $0.1, 0.23$ & $0.05, 1$ \\
\hline
Data noise 3 & $0.2, 0.5$ & $0.1, 2$ \\
\end{tabular} 
\end{table} 

\subsubsection{Linear conditioning results}\label{Res_2_1}
\begin{figure}
    \centering
    \includegraphics[width=.9\linewidth]{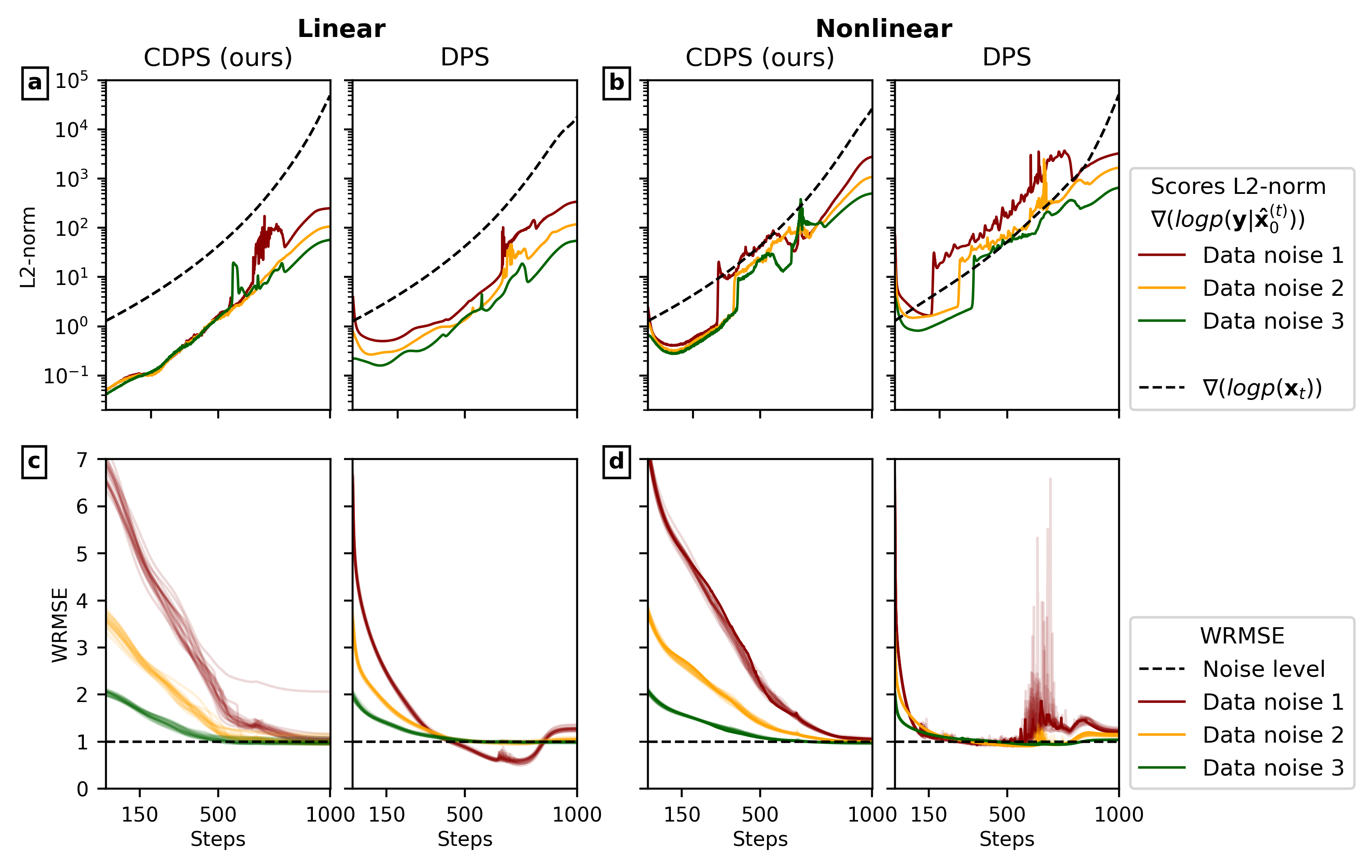}
    \caption{Evolution of prior (dashed) and likelihood (continuous) score functions generative steps, expressed as L2-norm, for (a) linear conditioning and (b) nonlinear conditioning case studies; and WRMSE of 100 samples generated as solutions of the (c) linear and (d) nonlinear inverse problems. The values of magnitude of the three data noise assumed are indicated in Table \ref{tab:Table2}.}
    \label{fig:inv_res_2}
\end{figure}

For the linear conditioning case, we considered three test cases involving the contamination of $\bf{d_{obs}}$ (Figure \ref{fig:inv_test_data}b) with different levels of uncorrelated Gaussian noise ($\sigma_{Fac}$ and $\sigma_{I_{P}}$, see Table \ref{tab:Table2}).  
Figure \ref{fig:inv_res_2}a shows the score magnitudes, expressed as L2-norms, for CDPS and DPS as a function of denoising/generative steps. For all settings, the scores of the data distribution have very similar values and increasing trends; an average for the three noise levels is plotted as dashed lines. The likelihood scores show generally an inverse correlation between their magnitude and the data noise level, highlighting a stronger constraining effect of low-noise $\bf{d_{obs}}$. This dependence is negligible for the first half of the denoising process for CDPS, with $ \nabla_{\textbf{x}_t}\log p_t(\textbf{d}|\textbf{x}_t)$ being around 30 times lower than $\nabla_{\textbf{x}_t} \log p_t(\textbf{x}_t)$. A similar ratio between prior and likelihood scores is observed in DPS for the the largest data noise case (\textit{Data noise 3}), while the overall likelihood score values generally increase with decreasing data error. At the end of the denoising process, the likelihood scores are very similar for both CDPS and DPS. This is expected as the two formulations become identical when the denoiser error becomes negligible.

For each scenario, we generated 100 conditional samples. The convergence of the CDPS process is shown in Figure \ref{fig:inv_res_2}c as the WRMSE between the predicted $\hat{\textbf{x}}_0^{(t)}$ ($\hat{\bf{x}}_0^{(t)} = \textbf{x}_0$ at $t=0$) and $\bf{d_{obs}}$. Between 93 and 95\% of the generated CDPS samples converged to a WRMSE $< 1.1$. For DPS, none of the samples converged when considering the lowest noise levels, and 99\% of the samples converged to a WRMSE <1.1 for the two larger-error cases. Overall, the WRMSE of the DPS samples approach lower values faster than CDPS during the denoising process; for \textit{Data noise 1}, the DPS samples ($\hat{\textbf{x}}_0^{(t)}$) initially overfit the data noise (WRMSE < 1) before diverging (WRMSE $\approx 1.5$). The WRMSE values obtained by CDPS also show comparatively larger variance (i.e., DPS samples follow similar paths).

\begin{figure}
    \centering
    \includegraphics[width=.9\linewidth]{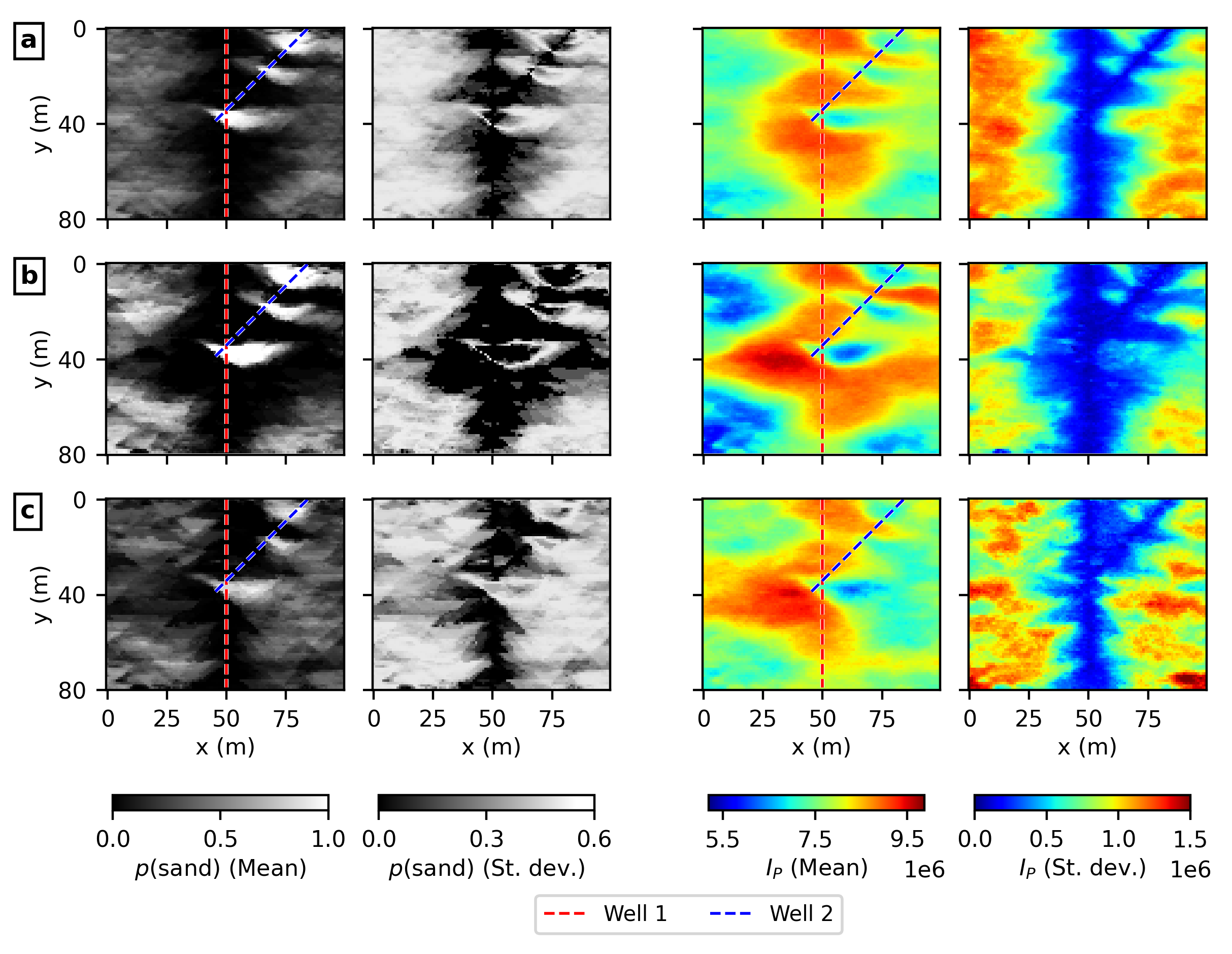}
    \caption{Point-wise average and standard distribution of Facies and $I_P$ posterior samples from (a) CDPS, (b) DPS, (c) pCN MCMC.}
    \label{fig:inv_res_2_1}
\end{figure}

Figures \ref{fig:inv_res_2_1}a and b show the ensembles of realizations sampled by CDPS and DPS, respectively, for the \textit{Data noise 2} case. The realizations generally converge correctly at well locations, and show increasingly larger variability with an increasing distance from the wells, approaching the prior model. Considering, as example, the unconditioned leftmost area between \textit{x} = 0 m and \textit{x} = 20 m, CDPS better represents this uniformity, with \textit{p}(sand) = $0.36$ and $I_P = (7.40 \pm 1.08) \times 10^6 \, \mathrm{Pa \cdot s / m}$, while DPS shows non-uniform channel distributions, with \textit{p}(sand) = $0.50$ and  $I_P = (7.12 \pm 0.88) \times 10^6 \, \mathrm{Pa \cdot s / m}$.
We further attempted to sample the posterior using pCN MCMC, considering $\beta$ = 0.03 and three independent chains, for 50000 iterations. After a burn-in time of 25000 iterations, only 60 parameters of the 120 conditioned parameters satisfied the $\hat{R}$ statistics threshold of 1.2. The ensemble of the sampled solutions are shown in Fig. \ref{fig:inv_res_2_1}c. Here, local conditioning is effective, but large variance (particularly evident in the facies distribution) highlight the points of non-convergence of the pCN. The unconditioned area between \textit{x} = 0 m and \textit{x} = 20 m has \textit{p}(sand) = $0.28$ and  $I_P = (7.90 \pm 1.03) \times 10^6 \, \mathrm{Pa \cdot s / m}$ and shows facies and $I_P$ spatial patterns closer to those of the CDPS. However, the samples show poor mixing: on average, the converged parameters show autocorrelation values lower than 0.2 only after 400 samples (i.e., each chain has approximately 62 independent samples).

\begin{figure}
    \centering
    \includegraphics[width=.5\linewidth]{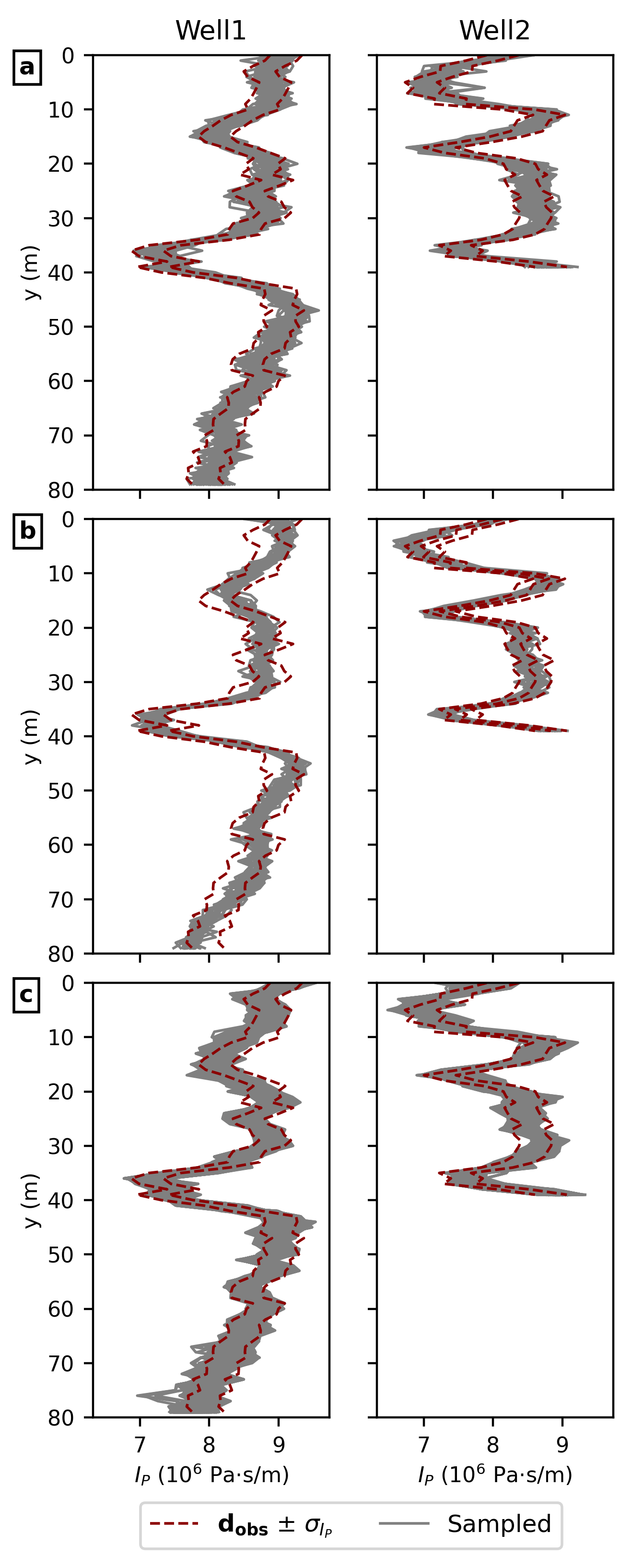}
    \caption{$I_P$ sampled distribution from (a) CDPS, (b) DPS, and (c) pCN MCMC, compared to the expected conditional data distribution $\bf{d_{obs}}$$\pm \sigma_{I_P}$.}
    \label{fig:inv_res_2_2}
\end{figure}

Figure \ref{fig:inv_res_2_2} shows the realizations of $I_P$ sampled by the three methods at the well locations, while their agreement with the target data ($\bf{d_{obs}}$ without noise) is summarized in Table \ref{tab:Table3}. The DPS results present a comparatively poorer fit to the data than CDPS (Fig. \ref{fig:inv_res_2_2}b), showing larger WRMSE and logS values than for CDPS. While the pCN results fit the data within the noise assumptions (Fig. \ref{fig:inv_res_2_2}c), it shows relatively larger logS  and RMSE values than CDPS to Well 2 data. These discrepancies may be due to the lack of sufficient posterior samples as indicated by the limited convergence of the MCMC chains. 

\begin{table}
\centering
\caption{RMSE and logS measures of the predicted $I_P$ for the linear inversion, considering the \textit{Data noise 2} case.}
\label{tab:Table3}
\begin{tabular}{ c || c | c | c || c | c | c }
\hline
     & \multicolumn{3}{c||}{Well 1} & \multicolumn{3}{c}{Well 2}  \\ 
\hline
     & CDPS & DPS & pCN & CDPS &  DPS & pCN \\ 
\hline
WRMSE & $0.70$ & $0.74$ & $0.72$& $0.79$ & $0.78$ & $0.86$ \\
\hline
LogS & $13.2$ & $14.0$ & $13.2$ & $13.3$ & $13.6$ & $13.7$\\
\hline
LogS (Prior) & \multicolumn{3}{c||}{15.9} & \multicolumn{3}{c}{15.7}  \\
\hline
\end{tabular} 
\end{table}

\subsubsection{Nonlinear inversion results}\label{Res_2_2}

We ran the nonlinear inversion example for three noise-levels in $\bf{d_{obs}}$ (Fig. \ref{fig:inv_test_data}c), considering absolute and relative components of the standard deviations ($\sigma_c$ and $r$, respectively, in Table \ref{tab:Table2}). 
Similar to the linear case, the prior scores (Fig. \ref{fig:inv_res_2}b) display an increasing trend that does not change significantly for different noise levels. The $ \nabla_{\textbf{x}_t}\log p_t(\textbf{d}|\textbf{x}_t)$ values of CDPS show a rather similar increasing trend for the three noise levels, ranging from one-tenth to similar values as $ \nabla_{\textbf{x}_t}\log p_t(\textbf{x}_t)$. Unlike the linear case, these scores show persistent differences between noise scales, with a maximum difference of one order of magnitude between the minimum and maximum data noise used. For DPS, the $ \nabla_{\textbf{x}_t}\log p_t(\textbf{d}|\textbf{x}_t)$ values  
are comparatively larger than CDPS and rapidly exceed the prior scores, becoming up to ten times larger (\textit{Data noise 1}). 

\begin{figure}
    \centering
    \includegraphics[width=1\linewidth]{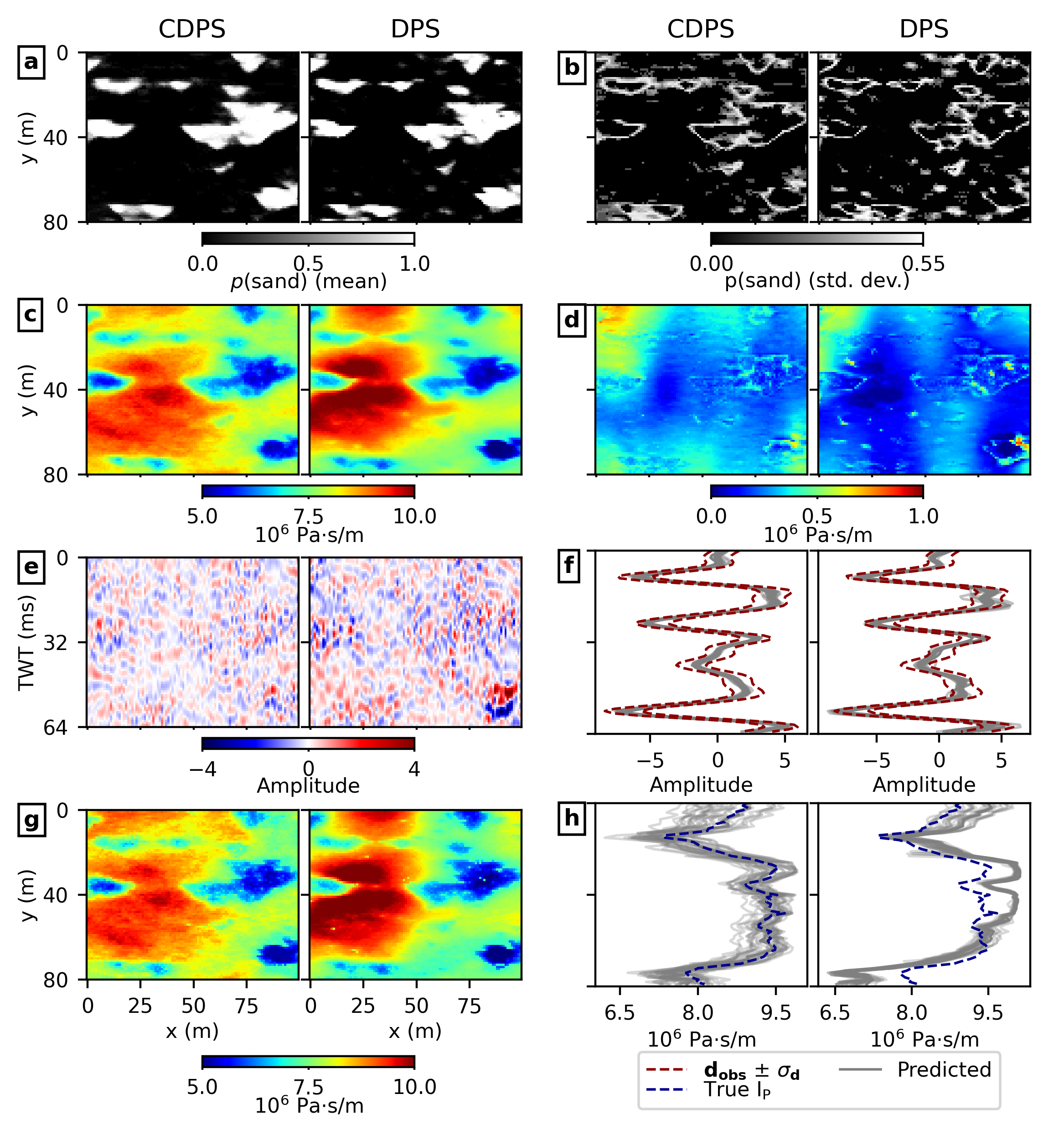}
    \caption{Sampled posterior distribution represented as (a) average facies; (b) facies standard deviation; (c) average $I_P$; (d) $I_P$ standard deviation; (e) average data error; (f) predicted data versus conditional data distribution ($\mathbf{d_{\text{obs}}} \pm \sigma_{\mathbf{d}}$) for a vertical trace at \textit{x} = 25 m; (g) $I_P$ realizations; (h) $I_P$ samples for a vertical trace at \textit{x} = 25 m compared to the unknown target distribution.}
    \label{fig:inv_res_2_3}
\end{figure}

All samples obtained with CDPS converged to a WRMSE $< 1.1$ (Fig. \ref{fig:inv_res_2}d). On the other hand, convergence of the DPS samples occurred only for the larger data noise case, while none of the samples converged for \textit{Data noise 1} (WRMSE = $1.23 \pm 0.03$) and \textit{2} (WRMSE = $1.13\pm0.02$). 
The results for \textit{Data noise 3} case are visualized in Fig. \ref{fig:inv_res_2_3} and corresponding metrics are shown in Table \ref{tab:Table4}. Comparing the seismic data computed from the posterior $I_P$ and the observed data without noise, the CDPS posterior presents comparatively larger accuracy (WRMSE$_{\bf{d}}$ in Table \ref{tab:Table4} and error distributions in Fig. \ref{fig:inv_res_2_3}e). However, both predictions have a forward response that is largely within the data error distributions (WRMSE$_{\bf{d}} < 1$). This is also shown by the trace plots in Figure \ref{fig:inv_res_2_3}f comparing predicted and conditioning data distributions for one vertical trace at \textit{x} = 25 m.

Overall, the CDPS shows large predictive power; the average facies and $I_P$ generated by CDPS (Fig. \ref{fig:inv_res_2_3}a and c) retrieve the target with high accuracy (lower RMSE$_\mathrm{F}$ and RMSE$_{I_P}$; Table \ref{tab:Table4}) and larger variability than DPS (Fig. \ref{fig:inv_res_2_3}b and d). The generated CDPS facies realizations predict the target facies with accuracies of 98\% and 95\%, showing a relative improvement compared to the DPS (97\% and 94\%, respectively), and having a large structural similarity (SSIM$_\mathrm{F}$ values in Table \ref{tab:Table4}).

The CDPS has superior prediction accuracy. For the whole study area, the generated ensemble captures the true $I_P$ much better than DPS, as shown by the lower values of logS in Table \ref{tab:Table4} and in the trace example in Figure \ref{fig:inv_res_2_3}h (\textit{x} = 25 m). Moreover, each realization reproduces the target $I_P$-distribution with a low $KL$ divergence (Table \ref{tab:Table4}) and presents fine-scale spatial variability patterns, which appear smoothed in DPS realizations (Fig. \ref{fig:inv_res_2_3}g).

In an attempt to compare the CDPS results with a high-quality posterior estimate, we ran the pCN method with $\beta$-values of 0.1, 0.05, 0.03, and 0.01, using three parallel chains. None of the runs converged to the posterior distribution after 50000 iterations, indicating that the inversion problem is hard, and implicitly suggesting that the CDPS method performs well in practice.

\begin{table}
\centering
\caption{Metrics of the posterior distributions sampled by CDPS and DPS for the non-linear inversion case, measured for data, facies and $I_P$ (subscripts $\bf{d}$, F, and $I_P$, respectively) for the \textit{Data noise 3} case.}
\label{tab:Table4}
\begin{tabular}{ c || c | c }
\hline
     & CDPS & DPS \\ 
\hline
WRMSE$\bf_{d}$ & $0.34\pm0.01$ & $0.39\pm0.01$ \\
\hline
RMSE$_\mathrm{F}$ & $0.16$ & $0.19$ \\
\hdashline
SSIM$_\mathrm{F}$ & $0.73\pm0.03$ & $0.67\pm0.01$ \\
\hline
RMSE$_{I_P}$ & $3.34 \times 10^5$ & $5.61 \times 10^5$\\
\hdashline
LogS$_{I_P}$ & $14.2$ & $36.8$ \\
\hdashline
$KL_{I_P}$ $(10^{-3})$ & $0.5\pm0.2$ & $1.3\pm0.2$ \\
\hline
\end{tabular} 
\end{table}

\subsection{Computational performance}\label{Res_3}
In this study, we did not consider any parallelization (see section \ref{codeimpl}). Moreover, we adopt a 2$^\mathrm{{nd}}$ order Heun solver when generating DM samples, and do not evaluate the costs/benefits of using a 1$^\mathrm{st}$ order solvers (e.g., Euler).
The U-Net adopted in the DM ($1.31\times10^8$ trainable parameters) required a total training time of 116 hours (600 epochs); the GAN ($1\times10^7$ parameters) and VAE ($3.2\times10^7$ parameters) took 21 and 10 hours, respectively, to train (2000 epochs). The DM exceeded the modeling accuracy of the fully trained GAN and VAE already at epoch 100, that is, after 19.5 hours of training (see Appendix \ref{AppA}).  

With EDM, accurate sampling could be obtained when using at least 18 diffusion steps, with no significant improvements observed when considering additional steps. The average generation time of the DM is 3 seconds per multivariate realization (5 to 7 times longer than for GAN and VAE). 
For the linear conditioning case study (Section \ref{Res_2_1}), the CDPS and DPS have similar computational time, 0.6 seconds per time step in our tests. The relative time increase, compared to unconditional modeling, is mostly due to the backpropagation computation for the likelihood scores. For the non-linear case (Section \ref{Res_2_2}), the CDPS had a larger computational cost, requiring 1.5 seconds per denoising step, mostly due to the additional Jacobian matrix computations. After testing different numbers of time-steps (from 18 to 2000 steps), we observed that both CDPS and DPS achieved their best modeling performances using 32 denoising steps. On the other hand, we observed significant differences when considering inversion accuracy for the non-linear case, as summarized in Table \ref{tab:Table5}. Here, the CDPS had a near-optimal performance using 250 steps, while the DPS required a minimum of 1000 steps, which, in our tests, resulted in a significant reduction of the sampling time (approximately half) when considering CDPS, despite its increased computational costs.

The time required by pCN MCMC for the linear case was substantially larger: the burn-in was reached after more than three days of computation, while the poor mixing of the sampled solutions show that much longer sampling would be necessary to recover the full posterior. We do not exclude that other hyperparameter values would improve the performance. The application of pCN on the nonlinear case did not converge in any of the considered cases.

\begin{table}
\centering
\caption{WRMSE$_{\bf{d}}$ for the three data noise cases considered in the nonlinear inverse modeling (Section \ref{Res_2_2}); "Div." (Diverged) indicates generative processes diverging to "Not a Number" (NaN) values.}
\label{tab:Table5}
\begin{tabular}{ c || c | c | c || c | c | c }
\hline
 & \multicolumn{3}{c ||}{CDPS} & \multicolumn{3}{c}{DPS} \\ 
Denoising steps & Data noise 1 & Data noise 2 & Data noise 3 & Data noise 1 & Data noise 2 & Data noise 3 \\ \hline
250 & 1.06\textpm0.03 & 1.01\textpm0.01 & 0.998\textpm0.005 & Div. & 1.30\textpm0.02 & 1.10\textpm0.02 \\
500 & 1.05\textpm0.02 & 0.99\textpm0.01 & 0.999\textpm0.005 & Div. & 1.18\textpm0.02 & 1.17\textpm0.02 \\
1000 & 1.05\textpm0.02 & 0.99\textpm0.01 & 0.997\textpm0.005 & 1.30\textpm0.03 & 1.17\textpm0.02 & 1.054\textpm0.004 \\
\hline
\end{tabular}
\end{table}

\section{Discussion}\label{Discussion}
The trained DM effectively samples from complex bivariate geological priors. To evaluate performance, we mainly focused on first- and second-order statistics (higher-order features are considered in the Appendix \ref{AppA}). The generated images match the TI features with a fidelity that, to our knowledge, is unprecedented for multivariate geological modeling with deep generative models. For the considered metrics, the DM outperformed the VAE and GAN implementations by \citet{mieleDeepGenerativeNetworks2024} (e.g., Fig. \ref{fig:mod_res_1} and Table \ref{tab:Table1}), except for a general overestimation of sand volume fraction that was slightly less pronounced in the VAE results (Table \ref{tab:Table1} and Figure S1 in Appendix \ref{AppA}). A practical advantage of DMs for multivariate modeling is the possibility to associate individual channels of the underlying U-Net to specific subsurface property types without the need to modify the network architecture. In contrast, VAE and GAN modeling required independent convolutional layers for each property type. This suggests that pre-designed DM architectures require none or only limited adjustments for modeling different geological scenarios.

The stability of the training process and the high accuracy of DM-based generative modeling come at the cost of high dimensionality and comparatively slow generation times. For instance, the benchmark VAE and GAN perform a compression from 16000 to 60 variables and the corresponding generation times are much faster \citep{mieleDeepGenerativeNetworks2024}. Modeling biases and errors occur, however, even when considering much lower compression rates, such as in LDMs \citep{ovangerStatisticalStudyLatent2025}, highlighting inherent trade-offs between speed, accuracy and dimensionality.

Using the proposed CDPS, pre-trained DMs can be flexibly modified to achieve noise-robust posterior sampling in both linear and non-linear inversion settings. The framework can be used for a wide range of inversion tasks without retraining, and it can deal with the simultaneous use of multiple conditioning data types (Appendix \ref{AppB}). The method includes modifications to the DPS by \citet{chung2023diffusion}, mainly in terms of an an improved likelihood score approximation. By accounting for the error in $\hat{\textbf{x}}_0^{(t)}$ on the simulated conditioning data (Eq. \ref{Eq: fixed_gradient_likelihood}), the likelihood scores become more accurate and scaled to diffusion noise (Fig. \ref{fig:inv_res_2}). With this modification, we obtain a much more stable generative process compared to DPS, balancing the two score contributions for posterior sampling (Eq. \ref{Eq: posterior_grad}) automatically (i.e., without arbitrary weighting), and making the likelihood score more independent of the data noise. However, a small noise-dependency remains for the nonlinear case with CDPS (Fig. \ref{fig:inv_res_2}b). This might be due to the overly simple assumption of homoscedastic noise in the $\hat{\textbf{x}}_0^{(t)}$ estimates (Fig. \ref{fig:err_sigmax0}), which may not hold if the trained network presents any form of localized bias (due to an incorrect training procedure, network's architecture and/or biased TI distributions). Improved error estimates may be needed in future applications, considering locally dependent variances or using approximations of Eq. \ref{Eq: TweediesVariance}.

Our proposed CDPS algorithm (Algorithm \ref{alg:Alg1}) enables efficient and more accurate local conditioning than DPS (see, e.g.,  Fig. \ref{fig:inv_res_2_1}a and b). We did not manage to properly benchmark the results against MCMC with pCN proposals (Fig. \ref{fig:inv_res_2_1}c). Indeed, even in the linear inversion case there is a relatively poor mixing and limited convergence of pCN MCMC; we did not investigate more advanced MCMC approaches such as parallel tempering formulations \citep{xuPreconditionedCrankNicolsonMarkov2020}. However, there is an overall good agreement between CDPS and pCN samples at the well locations (Fig. \ref{fig:inv_res_2_2} and Table \ref{tab:Table3}). This suggests that CDPS appropriately approximate the posterior. Compared with DPS (Fig. \ref{fig:inv_res_2_2}b), CDPS (Fig. \ref{fig:inv_res_2_2}a) has wider uncertainty bounds with magnitudes similar to those obtained by MCMC. We postulate that the underestimation by DPS is primarily a consequence of the likelihood score magnitudes being overestimated (Fig. \ref{fig:inv_res_2}a) when using Eq. \ref{Eq: xhat_likelihood}. For the non-linear inversion case, the CDPS performed well for all considered noise levels while DPS failed when considering low data noise levels (e.g., Fig. \ref{fig:inv_res_2} and Table \ref{tab:Table5}). Similarly to the linear case, we attribute the unstable DPS denoising trends (Fig. \ref{fig:inv_res_2}) to overestimated likelihood scores, with the overestimation increasing with decreasing data noise.

The DPS used herein is an implementation of the theory by \citet{chung2023diffusion} using the score-based EDM framework. We also considered arbitrarily-weighted Gaussian likelihood functions or the RMSE for the log-score approximation (as in the original DDIM implementation); this led to no convergence and unstable behavior in all the considered cases. We reproduce similar results using CDPS with DDIM (Appendix \ref{AppC}) highlighting that the method also works well for this formulation.

Even if the CDPS performs well in practice, it is still an approximate method because the logarithmic score estimate (Eq. \ref{Eq: fixed_gradient_likelihood}) relies on an approximation of the intractable likelihood (Eq. \ref{Eq: intract_likelihood}), the prior score estimates provided by the neural network will inevitably have some errors and the number of diffusion time steps are finite. One computational challenge of the CDPS method is the need to estimate the Jacobian of the forward operator at each time step. When this calculation is expensive, we anticipate that the CDPS approach would still work well in practice even if the Jacobian is only updated occasionally during the generative process. 

\section{Conclusions}
We investigate diffusion models (DM) for sampling multivariate geological priors and propose a modified conditional sampling method (CDPS) for approximate probabilistic Bayesian inversion. The method leverages the iterative nature of diffusion modeling, using an additional likelihood score approximation that modifies the generative process, allowing us to use any pre-traind, unconditional DM. The modeling performances are evaluated on a bivariate distribution of lenticular sand deposits and corresponding acoustic impedances ($I_P$). 
The unconditional DM simulations effectively captures first- and second-order statistics, as well as higher-order facies structures represented in training images (TI), consistently outperforming benchmark generative models based on VAEs and GANs. Bayesian posterior sampling is demonstrated for both linear (borehole data) and nonlinear (fullstack seismic) data scenarios, or combinations thereof. Compared to the original Diffusion Posterior Sampling (DPS) method \citep[DPS; ][]{chung2023diffusion}, our CDPS implementation provides stabilized inversion across noise levels, as well as more accurate posterior estimates, indicated by lower bias and enhancedr exploration of the posterior pdf. The method also shows significant advantages over an MCMC formulation in terms of convergence, leading to vastly decreased computing times. This work highlights the strengths of diffusion-based generative models for complex subsurface modeling and inference tasks. The training is much more robust than for GANs and the data generation quality is beyond the capabilities of GANs and VAEs. The CDPS approach enables flexible and scalable probabilistic inversion with minimal architectural tuning, resulting in an approximate Bayesian inversion method with broad applications across complex geological scenarios and data types.

\section{Code availability}
The implementation of CDPS for EDM is available at \url{https://github.com/romiele/CDPS_EDM}. The corresponding CDPS inversion using DDPM and DDIM frameworks are available at \url{https://github.com/romiele/CDPS_DDIM}. 

\bibliographystyle{unsrtnat}  

\bibliography{references}

\clearpage
\appendix

\section*{\LARGE Appendix}
\section{Additional assessment of prior modeling performance}\label{AppA}

\subsection{Geostatistical metrics} \label{A11}
\begin{figure}[h]
    \centering
    \includegraphics[width=.8\linewidth]{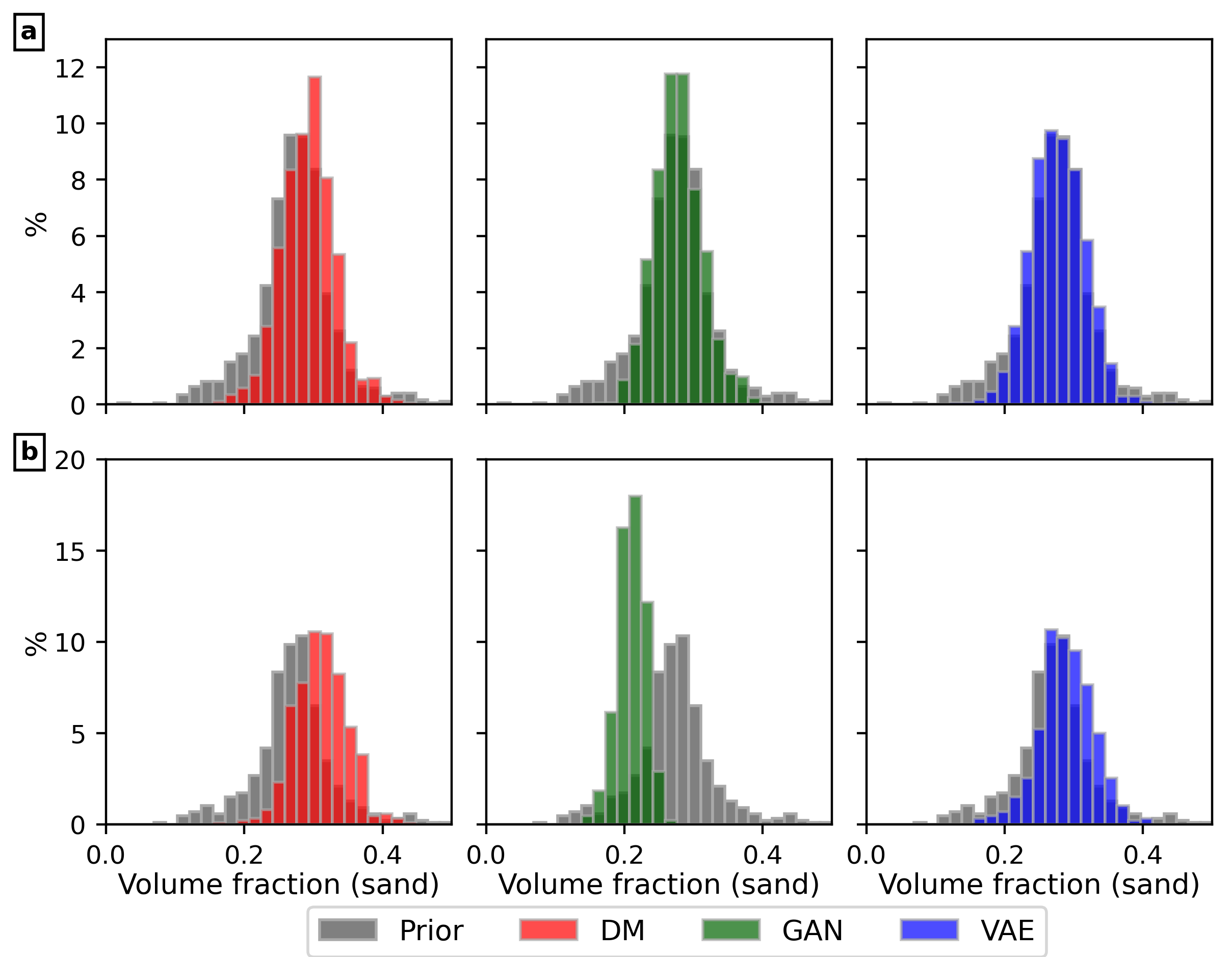}
    \caption{Volume fraction of sands in the generated realizations, compared to TI (Prior), for (a) the fully trained models, and (b) for early training stages at epoch 100.}
    \label{fig: A1_Volfrac}
\end{figure}

We further assessed the performance of unconditional modeling of the DM based on geological and geostatistical criteria. To evaluate the ability of the generative method to reproduce the channels' morphologies, Table \ref{tab:TableA1} summarizes the average and standard deviations of length, thickness and area of the sandy channels, as they are represented in the prior TI and reproduced by the DM, GAN and VAE. Here, the DM shows superior modeling ability, especially compared to the VAE where the unconditional modeling is expected to output unrealistic realizations \citep[see, e.g.,][]{levyVariationalBayesianInference2023, mieleDeepGenerativeNetworks2024, lopez-alvisDeepGenerativeModels2021}. We also observed that the realizations generated by the DM slightly overestimate the proportion of sands (0.29\textpm 0.04) compared to the targeted prior (0.27\textpm 0.05); VAE and GAN perform better (0.28\textpm 0.03 and 0.28\textpm 0.04, respectively) (Fig. \ref{fig: A1_Volfrac}a).

\begin{table}[H]
\centering
\caption{Measures for facies morphology computed from 500 prior/unconditional facies realizations.}
\label{tab:TableA1}
\begin{tabular}{ c | c | c | c }
\hline
     & Length (m) & Thickness (m) & Area (m$^2$) \\ 
\hline
Prior & 17.5 \textpm 4.3 & 5.9 \textpm 1.3 & 113 \textpm 33 \\
\hline
DM & 17.4 \textpm 4.4 & 5.9 \textpm 1.4 & 118 \textpm 36 \\
\hline
GAN & 16.2 \textpm 3.6 & 5.6 \textpm 1.2 & 105 \textpm 30 \\
\hline
VAE & 8.9 \textpm 1.6 & 3.3 \textpm 0.6 & 53 \textpm 13 \\
\hline
\end{tabular} 
\end{table}

\begin{figure}[h]
    \centering
    \includegraphics[width=.7\linewidth]{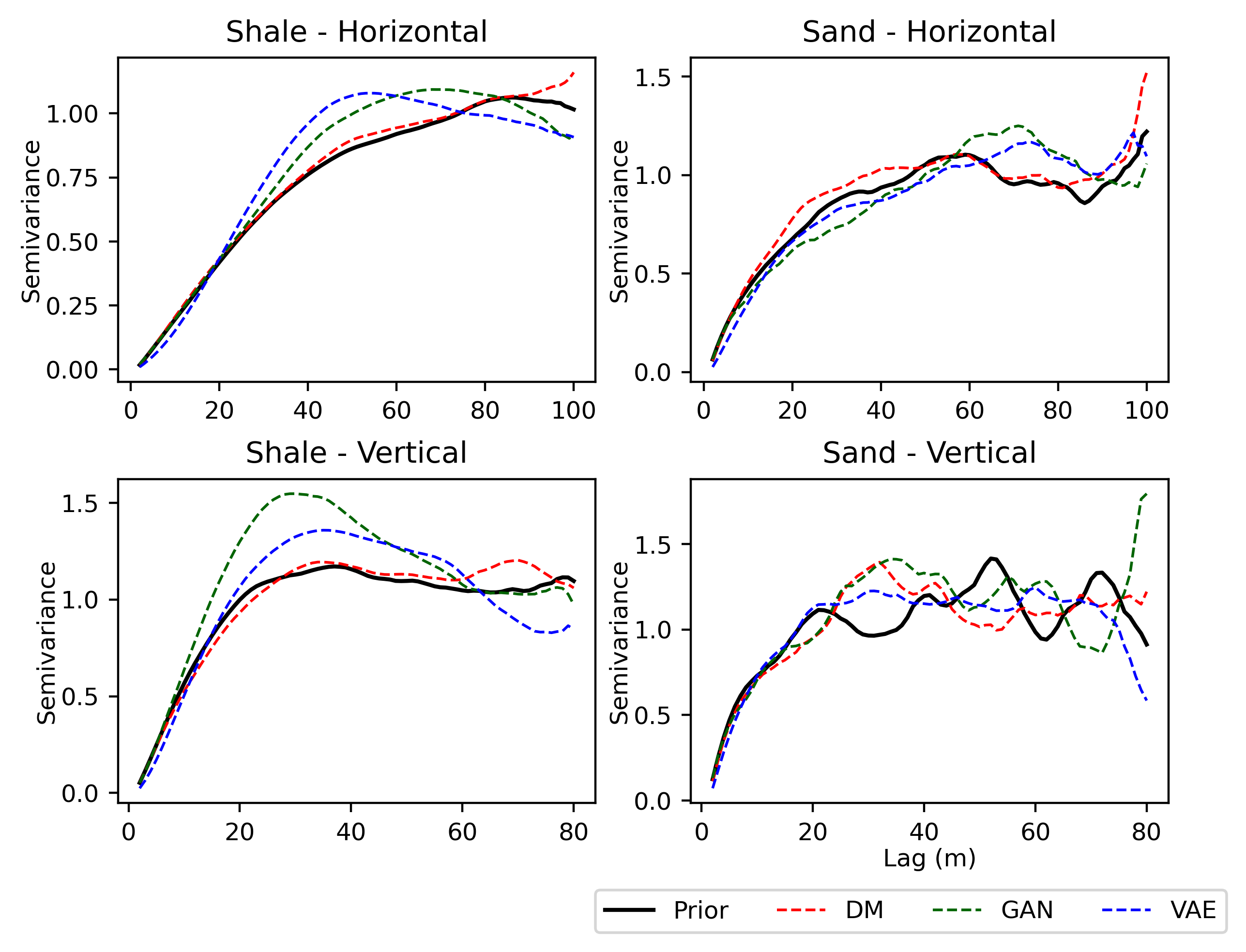}
    \caption{Experimental variograms, per facies and direction, computed from 60 $I_P$ prior and models realizations.}
    \label{fig: A2_Variograms}
\end{figure}

We further computed the spatial uncertainty patterns of the generated $I_P$ distributions as experimental variograms along the horizontal and vertical directions, from 60 realizations. The $I_P$ distributions in the TI were generated considering facies-dependent spatial patterns. For sands we used an exponential model, with maximum distance of significant spatial correlation \citep[ranges; e.g., ][]{deutschGSLIBGeostatisticalSoftware} of 50 m and 25 m along horizontal and vertical directions, respectively. For shales, we use a spherical variogram model with ranges of 65 m (horizontal direction) and 25 m (vertical direction). 
Due to the channels heterogeneity, the facies-dependent $I_P$ spatial patterns are affected by truncations at large ranges and their corresponding experimental variograms show disagreements with the theoretical models. However, we qualitatively estimate if spatial patterns are reproduced, by obtaining the experimental variograms from 100 TI samples and $I_P$ realizations generated with DM, GAN and VAE. The results are shown in Figure \ref{fig: A2_Variograms}, where the experimental points are linearly interpolated for visual aid. Overall, we see a good match between the DM and the prior, with noticeable improvements for the $I_P$ distribution in the shale.

\subsection{Training efficiency}\label{A12}
\begin{figure}[h]
    \centering
    \includegraphics[width=.6\linewidth]{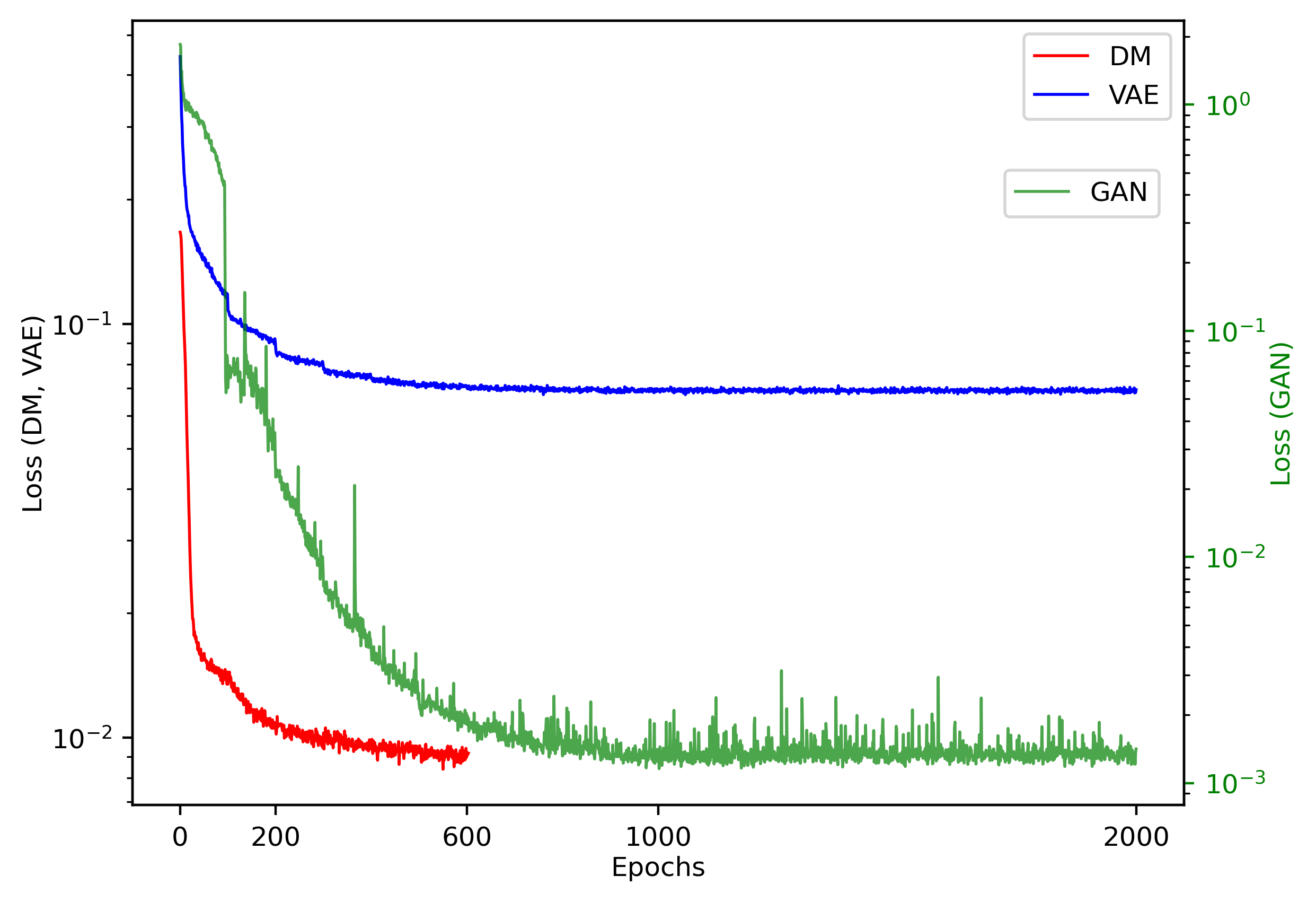}
    \caption{Loss values for the three trained networks. DM and VAE have comparable loss metrics (data mean squared error), GAN has an adversarial loss.}
    \label{fig: A2_epoch100}
\end{figure}

\begin{figure}[h]
    \centering
    \includegraphics[width=.9\linewidth]{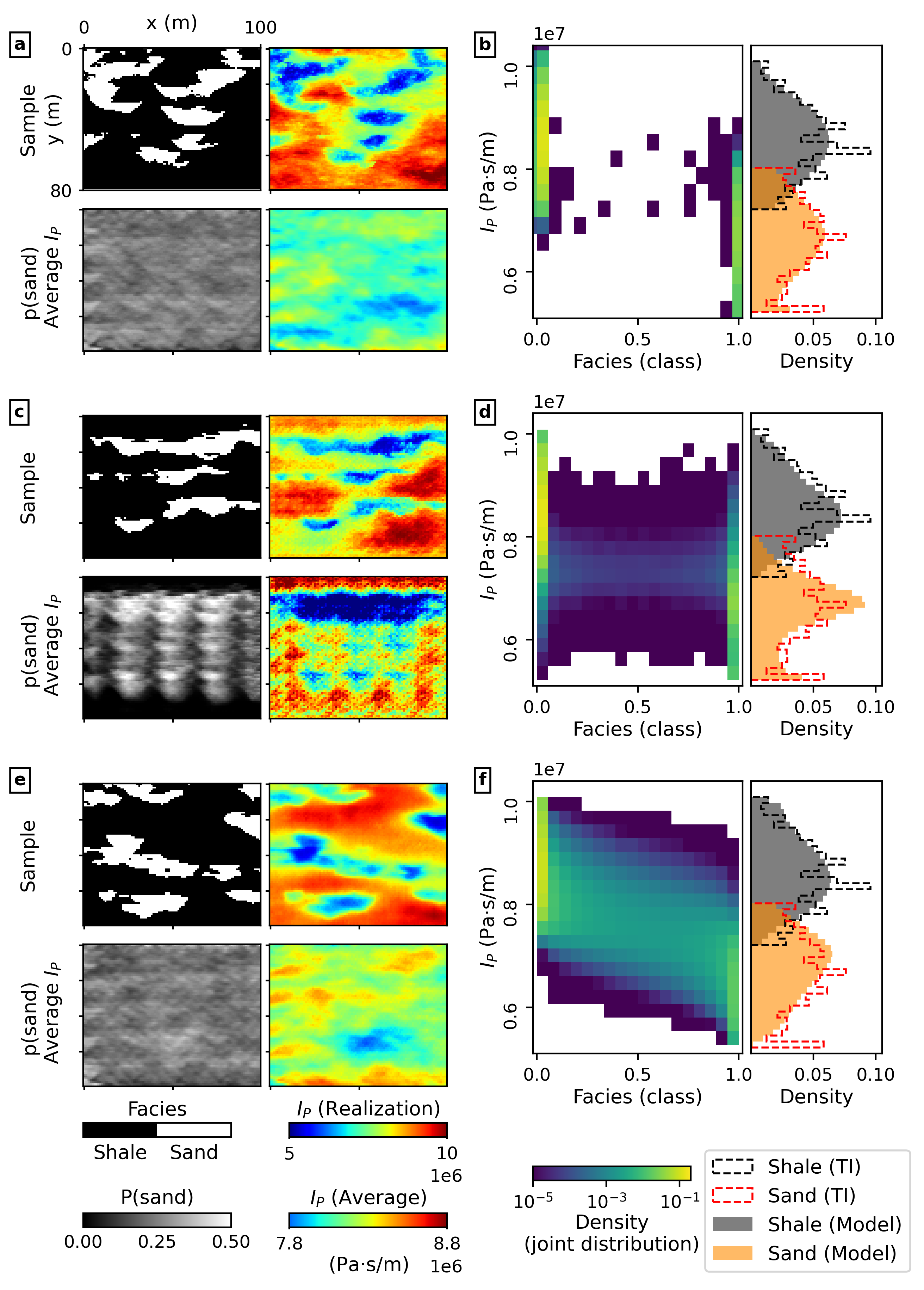}
    \caption{Summary of unconditional modeling performances of (a, b) DM, (c, d) GAN, and (e, f) VAE at training epoch 100. Subplots (a), (c), (e): realizations and average distribution of facies and $I_P$ for (a) DM, (c) GAN and (e) VAE; subplots (b), (d), (f): joint and $I_P$ distributions for (b) DM, (d) GAN and (f) VAE.}
    \label{fig: A3_epoch100}
\end{figure}

Diffusion models generally have a larger computational costs than GANs and VAEs. This is reflected, for example in the total training time of the proposed DM (4 days and 15h for 600 epochs) compared to the other generative models (respectively, 21h for the GAN and 12h for the VAE for 2000 epochs each). While we adopted a long training time for the three model to maximize their accuracy (Fig. \ref{fig: A2_epoch100}), we observe that the training of the DM is significantly more efficient. Already at epoch 100 (i.e., after 19h of training), the DM approximates the prior first- and second- order statistics with sufficient accuracy (Table \ref{tab:TableA2}, Fig. \ref{fig: A3_epoch100} and Fig. \ref{fig: A1_Volfrac}b) and it is generally more accurate than the fully trained GAN which required a similar training time. 

\begin{table}[h]
\centering
\caption{Metrics for unconditional prior modeling using the trained DM and benchmarking GAN and VAE at epoch 100. \textit{p}(sand) and Avg. $I_P$ refer to the average models obtained from the realizations ensembles. $KL$: Kullback–Leibler divergence; \textit{Avg.}: average; \textit{St. dev.}: standard deviation.}
\label{tab:TableA2}
\begin{tabular}{ c | c || c || c c }
\hline
     &  & \textit{p}(sand) \textemdash Avg. $I_P$ & \multicolumn{2}{c}{$I_P$ marginal distributions}\\ 
\hline
     & Lithology & Relative error & \begin{tabular}{@{}c@{}} {Avg.\textpm St. dev.} \\ {($10^6 \, \mathrm{Pa \cdot s / m}$)} \end{tabular}  &  $KL_{I_P}$\\ 
\hline
Prior & \begin{tabular}{@{}c@{}} \textit{Sand} \\ \textit{Shale} \end{tabular} & - & \begin{tabular}{@{}c@{}} 6.66 \textpm 0.73 \\ 8.54 \textpm  0.66\end{tabular} & - \\
\hline
DM    & \begin{tabular}{@{}c@{}} \textit{Sand} \\ \textit{Shale} \end{tabular} & 
        \begin{tabular}{@{}c@{}} 15.0\% \\ 0.82\% \end{tabular} & 
        \begin{tabular}{@{}c@{}} 0.665 \textpm 0.73 \\ 0.853 \textpm  0.67 \end{tabular} & 
        \begin{tabular}{@{}c@{}} 0.8e-3 \\ 0.9e-3 \end{tabular} \\
\hline
GAN   & \begin{tabular}{@{}c@{}} \textit{Sand} \\ \textit{Shale} \end{tabular} & 
        \begin{tabular}{@{}c@{}} 58.0\% \\ 3.8\% \end{tabular} &
        \begin{tabular}{@{}c@{}} 0.662 \textpm  0.65 \\ 0.853 \textpm 0.63\end{tabular} & \begin{tabular}{@{}c@{}} 2.3e-3 \\ 1.4e-3 \end{tabular} \\
\hline
VAE  & \begin{tabular}{@{}c@{}} \textit{Sand} \\ \textit{Shale} \end{tabular} & 
        \begin{tabular}{@{}c@{}} 12.3\% \\ 1.2\% \end{tabular} &
        \begin{tabular}{@{}c@{}} 0.686 \textpm  0.70\\ 0.862 \textpm 0.68\end{tabular} & 
        \begin{tabular}{@{}c@{}} 1.4e-3 \\ 1.8e-3 \end{tabular} \\
\hline
\end{tabular} 
\end{table}

\clearpage
\section{Conditioning on multiple data types}\label{AppB}

\begin{figure}[h] 
    \centering
    \includegraphics[width=.9\linewidth]{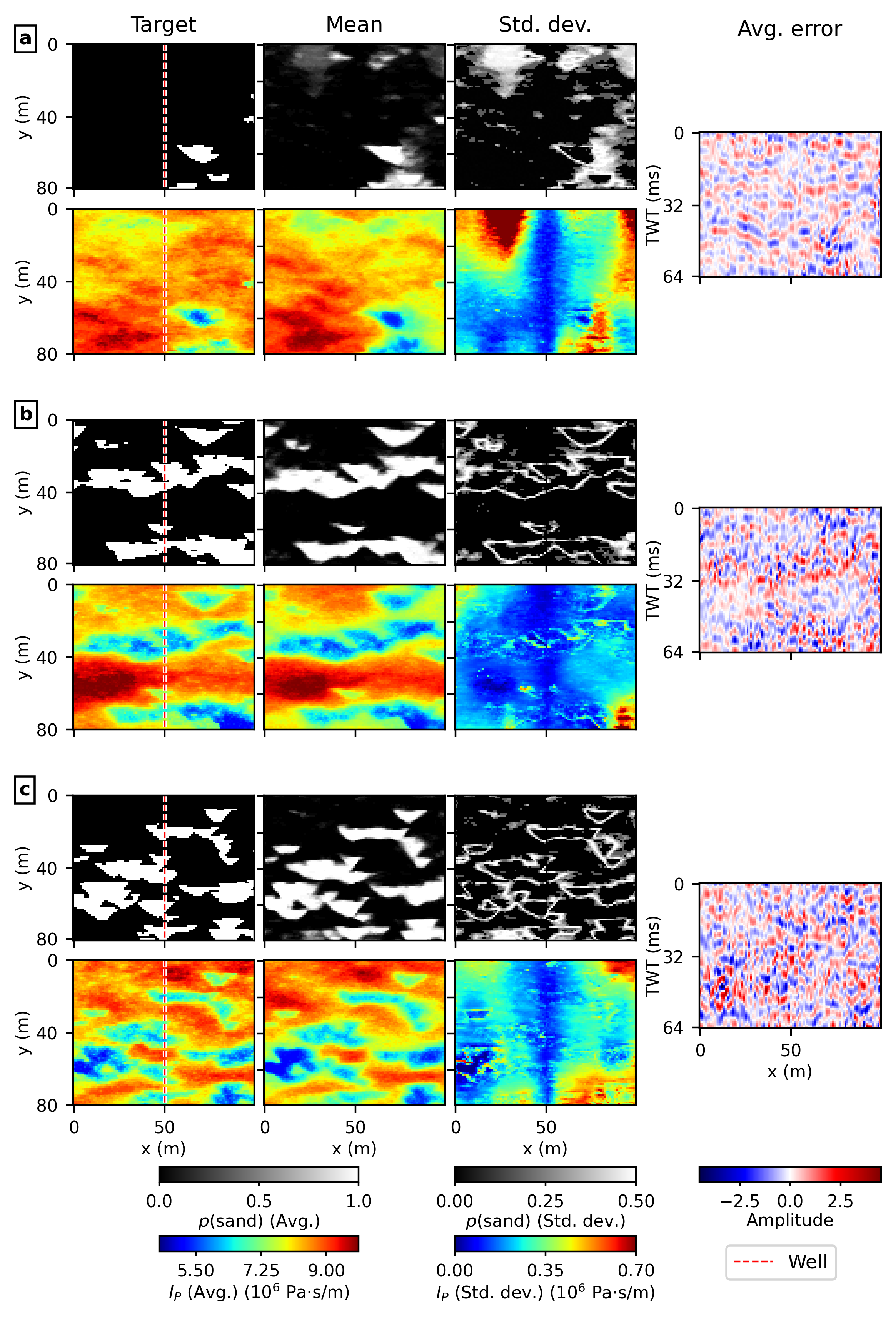}
    \caption{Target and posterior facies and $I_P$ for the three inversion case studies (a, b, c) using both well logs and seismic data as conditioning data. The sampled posterior distribution is shown as mean and standard deviation (\textit{Std. dev.}) of realizations' ensembles, and corresponding average seismic data misfit.}
    \label{fig: B1}
\end{figure}

We applied the CDPS algorithm on three different test cases (\textit{Target} in Fig. \ref{fig: B1}a, b, c) using both direct observations (well logs) and seismic reflection data; the location of the well is shown in Fig. \ref{fig: B1}
as red dashed line. The conditioning data has Gaussian uncorrelated noise with amplitude corresponding to the \textit{Data noise 2} (Table 2 in Section 3.2). Conditioning on two data types required computing the corresponding likelihood scores, individually, and summing them to the prior score for the denoising at each time-step. For these applications we considered 300 steps. The target used as \textit{Case 1} (Fig. \ref{fig: B1}a) represents a scenario not included in our TI, having only a single channel body in the facies distribution (2\% in sand volume fraction). This can be seen as an example of inversion when the prior knowledge is biased; \textit{Case 2} and \textit{Case 3} (Fig. \ref{fig: B1}b and c, respectively) are test example extracted from the TI (not used during training). 

\begin{figure}[h]
    \centering
    \includegraphics[width=.9\linewidth]{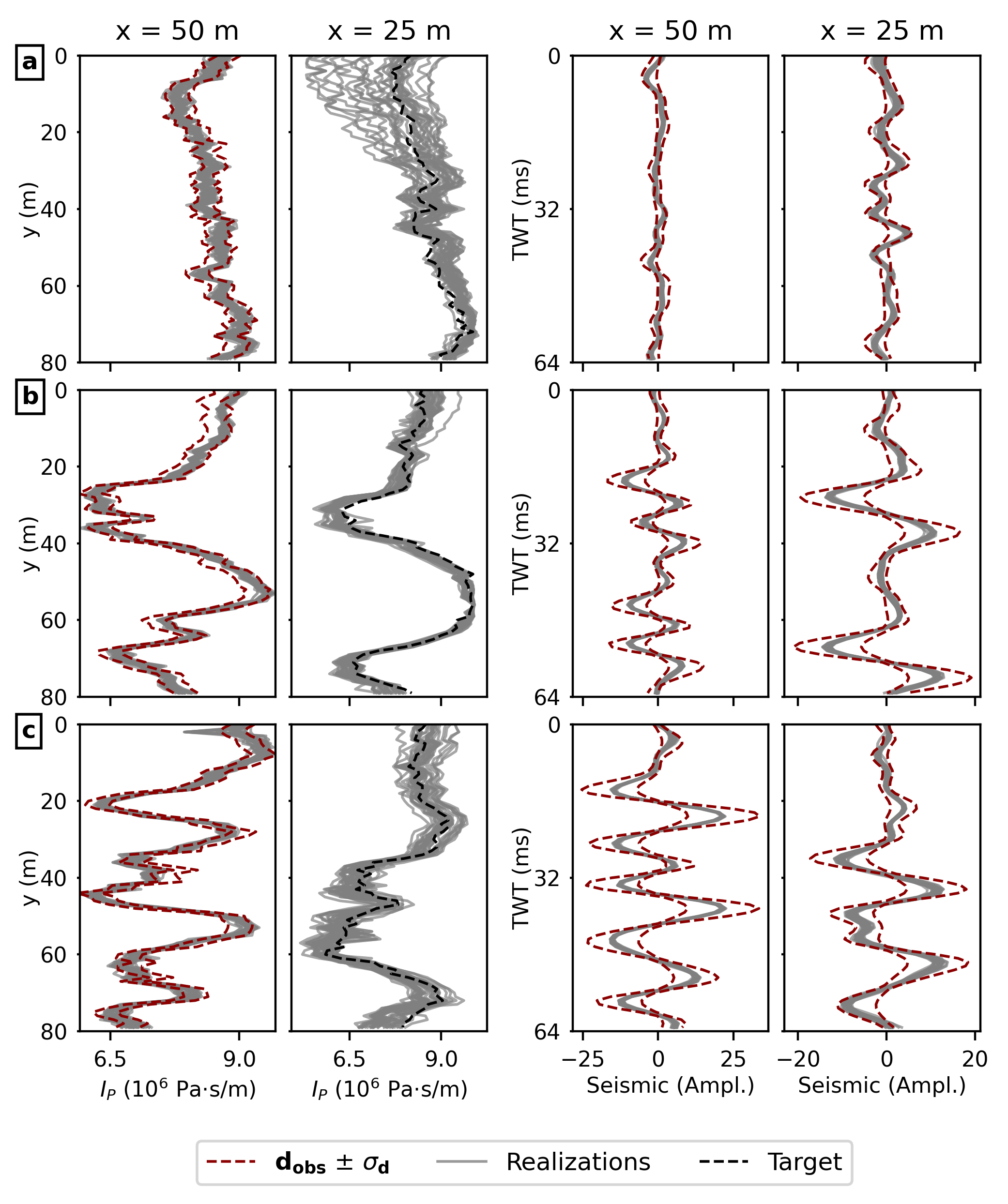}
    \caption{Predicted $I_P$ and corresponding seismic data along vertical traces located at the well location (x = 50 m) and at x = 25 m, for the three inversion case studies (a, b, c). The predictions are compared with conditioning data $\bf{d_{obs}}$ and unknown true $I_P$ (\textit{Target}).}
    \label{fig: B2}
\end{figure}

Figures \ref{fig: B1} and \ref{fig: B2} summarize the solutions obtained, showing the results from 50 generated samples; the accuracy of the inversion solutions is described in Table \ref{tab:TableB1} using the same metrics shown in Section 3.2.2. In all the cases, we observe a good convergence of the solutions, with all the samples having WRMSE (when compared to the noiseless data) $<1$ (WRMSE$_\textbf{d}$ in Table \ref{tab:TableB1}). The solutions retrieved honor the conditioning data within the uncertainty ranges, in both well data and seismic data domains (Fig. \ref{fig: B2}): at the well location (x = 50 m in Fig. \ref{fig: B2}), the $I_P$ values fit the uncertainty range, while the corresponding seismic data shows lower variability than the considered error standard deviation, especially at larger seismic amplitudes. A similar pattern is observed at locations where only seismic data conditioning is used (e.g.,  x = 25 m in Fig. \ref{fig: B2}). Given the close match between the predicted and noiseless seismic data, this behavior suggests that the relative component of the seismic uncertainty does not significantly impact the inversion results.

Both facies and $I_P$ distributions match the target with good accuracy in the studied area (Fig. \ref{fig: B2}). Overall, the facies realizations generated match the target with an accuracy of 94\% for \textit{Case 1} and 96\% for \textit{Case 2} and \textit{Case 3}, with a large structural similarity to the target (SSIM$_\mathrm{F}$ in  Table \ref{tab:TableB1}).  The \textit{p}(sand) and average $I_P$ computed from the realizations ensembles (Fig. \ref{fig: B1}) match the targets, and their corresponding standard deviation shows larger variability mostly at facies interfaces. The standard deviation of $I_P$ realizations  (\textit{Std. dev.} in Fig. \ref{fig: B1}) also highlights the lateral conditioning of the well log data. For \textit{Case 1}, few solutions predict the presence of sands in the upper left area (an area of lower $I_P$ values), significantly mismatching the target; here, we also see larger variability of the solutions (Fig. \ref{fig: B1} and, e.g., $I_P$ trace at x = 25 m in Fig. \ref{fig: B2}a). This can be interpreted as an effect of the bias of the trained DM compared to this case study. 

Overall, the predictive power of the CDPS is large, with good results for the biased \textit{Case 1}, as shown by the comparably close values of all the metrics in Table \ref{tab:TableB1}. The \textit{Case 2} shows relatively poorer performances of the CDPS in $I_P$ distribution prediction. The realizations present lower variability (Figures \ref{fig: B1}b and \ref{fig: B2}b), and their ensemble not always match well to the local true $I_P$, hence the higher logS measured values (Table \ref{tab:TableB1}). 

\begin{table}[h]
\centering
\caption{Metrics of the posterior distributions sampled by CDPS, measured for data, facies and $I_P$ (subscripts $\bf{d}$, F, and $I_P$, respectively), for the three studied cases assuming \textit{Data noise 2}.}
\label{tab:TableB1}
\begin{tabular}{ c  ||c  c c}
\hline
     & Case 1& Case 2 &Case 3\\ 
\hline
WRMSE$\bf_{d}$ & 0.204\textpm0.005& 0.199\textpm0.006&0.201\textpm0.06\\
\hline
RMSE$_\mathrm{F}$ & 0.19& 0.20&0.21\\
\hdashline
SSIM$_\mathrm{F}$ & 0.73\textpm0.05& 0.72\textpm0.02&0.69\textpm0.02\\
\hline
RMSE$_{I_P}$ & 4.50 $\times 10^5$ & 3.58 $\times 10^5$ & 3.24 $\times 10^5$\\
\hdashline
LogS$_{I_P}$ & 14.5 & 18.7 & 14.2\\
\hdashline
$KL_{I_P}$ $(10^{-4})$& 6.3\textpm1.5& 3.5\textpm0.5&3.8\textpm0.8\\
\hline
\end{tabular} 
\end{table}


\clearpage
\section{CDPS implementation with DDPM and DDIM}\label{AppC}
We implement the CDPS in the DDPM and DDIM frameworks, shown in Algorithm \ref{alg:Alg}. Differently from score-based diffusion, the trained network predicts the noise added to a sample, denoted as $\epsilon_\theta$. This can be related to the score function using the following correlation \citep{dhariwalDiffusionModelsBeat2021} 
\begin{equation}\label{Eq: C1}
    \nabla_{x_t} \log p_t (\textbf{x}_t) = -\frac{1}{\sqrt{1-\bar{\alpha}_t}}\epsilon_\theta^{(t)},
\end{equation}

where $\bar{\alpha}_t$ is inversely correlated to the diffusion noise (See Section 2.1 and \citet{hoDenoisingDiffusionProbabilistic2020}).  Therefore, to integrate the likelihood score within the DDPM and DDIM frameworks, we rescale it to $\sqrt{(1-\bar{\alpha}_t)} \nabla_{x_t} \log p_t (\textbf{d}|\textbf{x}_t)$ (line 19 of Algorithm \ref{alg:Alg}). We then sum the two components as
\begin{equation}\label{Eq: C2}
   \hat{\epsilon}^{(t)} = \epsilon_\theta^{(t)} - \sqrt{(1-\bar{\alpha}_t)} \nabla_{x_t} \log p_t (\textbf{d}|\textbf{x}_t),
\end{equation}

and use this "posterior-related" drift to obtain the denoised $\textbf{x}_{t-1}$. This approach was first proposed by \citet{dhariwalDiffusionModelsBeat2021} in their guided diffusion modeling. For the application of the CDPS with DDIM, we use the same diffusion UNet described in Section 2.3 and retrain it using the same TI for noise prediction; the training parameters and process are the same described in  \citet{dhariwalDiffusionModelsBeat2021}. 

\begin{figure}[h]
    \centering
    \includegraphics[width=.7\linewidth]{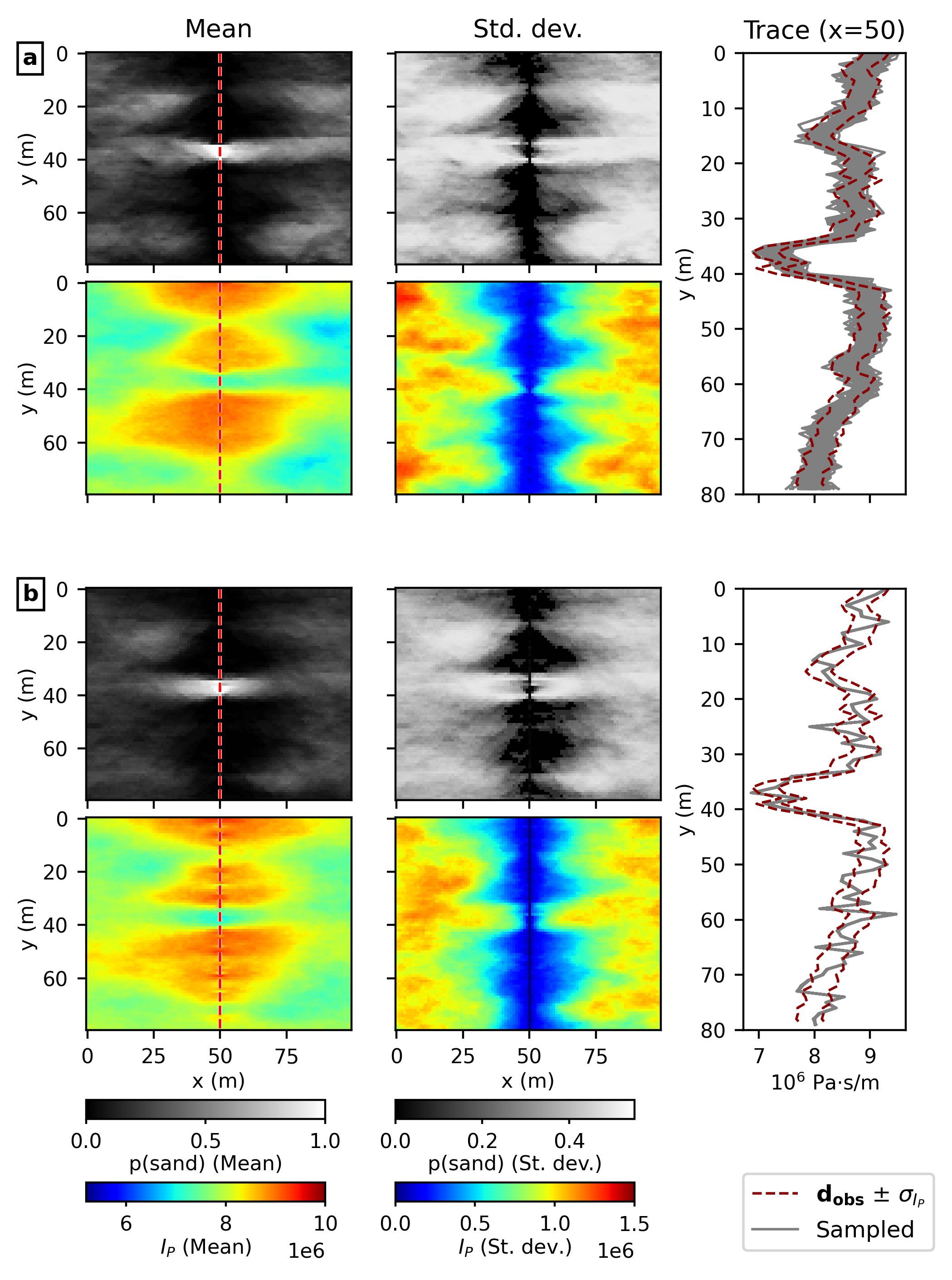}
    \caption{Conditioned realizations sampled using (a) CDPS and (b) DPS, represented as mean and standard deviation (\textit{Std. dev.}) of facies and $I_P$ distributions, and sampled $I_P$ at the conditioning well location (x=50m).}
    \label{fig: C1_trace}
\end{figure}

Using one well as conditioning data, contaminated with Gaussian noise of magnitude equal to the case \textit{Data noise 2} (Table 2 in Section 3.2), we sampled 100 samples using CDPS (Algorithm \ref{alg:Alg}) and the DPS method in its original implementation, with weight $\rho = 1$ as defined for the inpainting tasks. Using CDPS with either DDIM (deterministic) or DDPM (stochastic) sampling did not change the results; we used the DPS in its original implementation with DDPM. For both CDPS and DPS, we sampled with 1000 denoising steps. 
The results are summarized in Fig.  \ref{fig: C1_trace}. All the realizations using CDPS match the conditioning facies with 100\% of accuracy and have WRMSE = 0.7 \textpm 0.2; in the unconditioned area between x = 0 m and x = 20 m, the \textit{p}(sand) = 0.28. These values are closer to the prior than our linear conditioning case with score-based diffusion (Section 3.2.1); we attribute this difference to the absence of the second conditioning well. 
For DPS, we observe that the data is perfectly matching the noisy data (overfitting), with standard deviation approaching 0. This does not occur in our implementation of the DPS with score-based diffusion, as the likelihood score magnitude is rescaled proportionally to the diffusion noise content, hence is less \textit{influential} on the conditioning diffusion. The lack of such rescaling is, in fact, the inconsistency between conditional diffusion theory and the original DPS implementation on DDPM we highlight in Section 2.2.1. Our CDPS (Algorithm \ref{alg:Alg}) integrates this correction. 

\begin{algorithm}
    \caption{Corrected DPS algorithm for DDPM (Gaussian error)}\label{alg:Alg}
    \begin{algorithmic}[1] 
    
    \State \textbf{Require:} trained network $S_\theta(\textbf{x}_t, t)$, conditioning data $\textbf{d}$, forward operator $\mathcal{F}$, noise schedule $\{\sigma\}_{t=1}^{N}$ and $\{\beta\}_{t=1}^{N}$ (with $\alpha=1-\beta$) 
    \newline
        
    \For{ \textit{t = T,..., 1}} \Comment{Evaluate the denoiser uncertainty at sampling noise levels}
        \State \textbf{sample} $\textbf{x}_0 \sim q_{data}$
        \State $\textbf{x}_t \gets \alpha_t \textbf{x}_0 + (1-\alpha_t) \textbf{z}, \quad \textbf{z} \sim \mathcal{N}(0,I)$
        \State $\epsilon_\theta^{(t)}  \gets S_\theta(\textbf{x}_t, t)$
        \State $\hat{\textbf{x}}_0^{(t)} \gets \left(\textbf{x}_t -\sqrt{1-\bar{\alpha}_t}  \epsilon_\theta^{(t)}\right)/ \sqrt{\bar{\alpha}_t}$ 
        \State $\sigma_{{\hat{x}_0}}^{(t)}\gets \mathbb{E}\left[\sqrt{\sum_{i=1}^N(\hat{x}_{0,i}^{(t)}-x_{0,i})^2/N}\right]$
    \EndFor
    \Statex
    \State \textbf{sample} $\textbf{x}_T \sim \mathcal{N}(0,I)$
    \For{\textit{t = T-1,..., 0}}
        \State $\epsilon_\theta^{(t)} \gets S_\theta(\textbf{x}_t, t)$ \Comment{Evaluate the added noise at $t$}
        \State $\hat{\textbf{x}}_0^{(t)} \gets \left(\textbf{x}_t -\sqrt{1-\bar{\alpha}_t}  \epsilon_\theta \right)/ \sqrt{\bar{\alpha}_t}$  \Comment{Evaluate the denoised image at $t$}
    \Statex
    \State $\Sigma_{\hat{\textbf{x}}_0}^{(t)} \gets \sigma_{\hat{\textbf{x}}_0}^{2^{(t)}}$  \Comment{Get denoiser uncertainty at $t$}
    \If{$\mathcal{F}$ is nonlinear} \Comment{Propagate denoiser uncertainty in the \textbf{d} domain}
        \State $\mathcal{J}_{\hat{\textbf{x}}_0} \gets \mathcal{J}_\mathcal{F} (\hat{\textbf{x}}_0^{(t)})$
        \State $\tilde{\Sigma}_\textbf{d} \gets \mathcal{J}_{\hat{\textbf{x}}_0^{(t)}}^T \Sigma_{\hat{\textbf{x}}_0}^{(t)} \mathcal{J}_{\hat{\textbf{x}}_0^{(t)}} + \Sigma_\textbf{d}$
    \Else
        \State $\tilde{\Sigma}_\textbf{d} \gets \mathcal{F}^T \Sigma_{\hat{\textbf{x}}_0}^{(t)} \mathcal{F} + \Sigma_\textbf{d}$
    \EndIf
    \Statex
    \State $\nabla_{x_t} \log p_t (\textbf{d}|\hat{\textbf{x}}_0^{(t)}) \gets -\nabla_{\textbf{x}_t} \left[(\mathcal{F}(\hat{\textbf{x}}_0^{(t)})-\textbf{d})^T \tilde{\Sigma}_\textbf{d}^{-1} (\mathcal{F}(\hat{\textbf{x}}_0^{(t)})-\textbf{d})\right]$ \Comment{Evaluate likelihood score at $t$}
    \State $\hat{\epsilon}^{(t)}  \gets  \epsilon_\theta^{(t)} + \sqrt{(1-\bar{\alpha}_t)} \nabla_{x_t} \log p_t (\textbf{d}|\hat{\textbf{x}}_0^{(t)})$  \Comment{Evaluate "posterior" noise at $t$}
    \Statex
    \State $\textbf{z} \sim \mathcal{N}(0,I)$
    \State $x_{t-1} \gets \frac{\sqrt{\alpha_t}(1-\bar{\alpha}_{t-1})}{1-\bar{\alpha}_{t-1}}\textbf{x}_t + \frac{\sqrt{\bar\alpha_{t-1}}\beta_t}{1-\bar{\alpha}_{t-1}} \left(\textbf{x}_t -\sqrt{1-\bar{\alpha}_t}  \nabla_{x_t} \log p_t (\hat{\textbf{x}}_0^{(t)}|\textbf{d})  \right)/ \sqrt{\bar{\alpha}_t} + \tilde{\sigma}_t \textbf{z}$ \Comment{Take Euler step}

\EndFor
\State \textbf{Output:} $\tilde{\textbf{x}}_0 \sim p(\textbf{x}_0|\textbf{d})$
\end{algorithmic} 
\end{algorithm}

\begin{figure}[h]
    \centering
    \includegraphics[width=.8\linewidth]{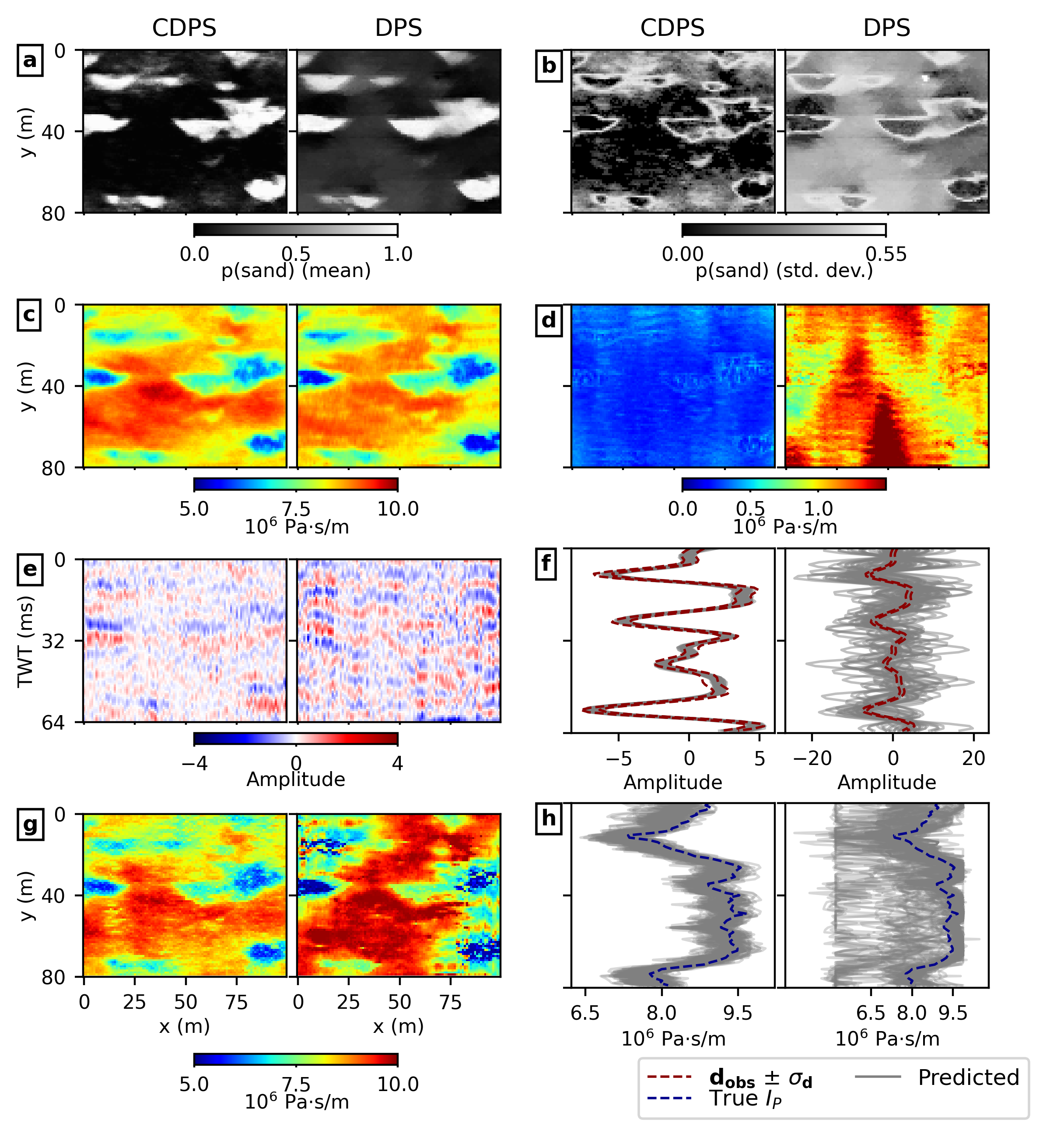}
    \caption{Sampled posterior distribution represented as (a) average facies; (b) facies standard deviation; (c) average $I_P$; (d) $I_P$ standard deviation; (e) average data error; (f) predicted data versus conditional data distribution ($\mathbf{d_{\text{obs}}} \pm \sigma_{\mathbf{d}}$) for a vertical trace at x = 50m; (g) $I_P$ realizations; (h) $I_P$ samples for a vertical trace at x = 25m compared to the unknown target distribution.}
    \label{fig: C2_seis}
\end{figure}

We further applied the methods for fullstack seismic data inverse modeling. We show here the results for a nonlinear inverse problem equal to that use in Section 3.2.2, considering the \textit{data noise 2}. For DPS, we show the results considering a likelihood score weight $\rho=0.4$, corresponding to that suggested by \citet{chung2023diffusion} for nonlinear phase retrieval.  For both CDPS and DPS, we sampled with 1000 denoising steps.
The results are summarized in Fig. \ref{fig: C2_seis} and integrated with the metrics of Table \ref{tab:TableC1}. The DPS showed high instability of the denoising process; the CDPS underperformed compared to the equivalent score-based inverse problem. Although we did not investigate the possible causes for such difference, we believe that assumptions of an isotropic uncertainty for the denoiser may play a role in the inversion accuracy for the nonlinear case. 

\begin{table}[h]
\centering
\caption{Metrics of the posterior distributions sampled by CDPS and DPS, measured for data, facies and $I_P$ (subscripts $\bf{d}$, F, and $I_P$, respectively).}
\label{tab:TableC1}
\begin{tabular}{ c || c | c }
\hline
     & CDPS & DPS \\ 
\hline
WRMSE$\bf_{d}$ & 0.70\textpm0.01& 7.1\textpm0.6\\
\hline
RMSE$_\mathrm{F}$ & 0.19& 0.24\\
\hdashline
SSIM$_\mathrm{F}$ & 0.60\textpm0.04& 0.51\textpm0.15\\
\hline
RMSE$_{I_P}$ & 4.47 $\times 10^5$& 4.19 $\times 10^5$\\
\hdashline
LogS$_{I_P}$ & 14.0 & 16.0\\
\hdashline
$KL_{I_P}$ $(10^{-3})$ & 0.4\textpm0.1& 3.1\textpm1.1\\
\hline
\end{tabular} 
\end{table}



\end{document}